\documentclass[preprint,12pt]{elsarticle}

\usepackage{amssymb}
\usepackage{amsmath}

\usepackage{graphicx}%
\usepackage{multirow}%
\usepackage{amsmath,amssymb,amsfonts}%
\usepackage{amsthm}%
\usepackage{mathrsfs}%
\usepackage[title]{appendix}%
\usepackage{xcolor}%
\usepackage{textcomp}%
\usepackage{manyfoot}%
\usepackage{booktabs}%
\usepackage{algorithm}%
\usepackage{algorithmicx}%
\usepackage{algpseudocode}%
\usepackage{listings}%

\usepackage{subcaption}
\usepackage{comment}
\usepackage{colortbl}
\usepackage{array}

\usepackage{makecell}
\usepackage{adjustbox}
\usepackage{float}

\usepackage[none]{hyphenat}
\usepackage{caption}

\usepackage{hyperref}
\usepackage[percent]{overpic}

\journal{Image and Vision Computing}

\begin{document}

\begin{frontmatter}



\title{NV3D: Leveraging Spatial Shape Through Normal Vector-based 3D Object Detection}




\author[siit]{Krittin Chaowakarn}
\author[nectec]{Paramin Sangwongngam}
\author[chula]{Nang Htet Htet Aung}
\author[siit]{Chalie Charoenlarpnopparut}

\affiliation[siit]{organization={The School of Information, Computer, and Communication Technology, 
            Sirindhorn International Institute of Technology, Thammasat University},
            city={Pathum Thani},
            postcode={12120},
            country={Thailand}}

\affiliation[nectec]{organization={National Electronics and Computer Technology Center, 
            National Science and Technology Development Agency},
            city={Pathum Thani},
            postcode={12120},
            country={Thailand}}

\affiliation[chula]{organization={Department of Electrical Engineering, Faculty of Engineering,
            Chulalongkorn University},
            city={Bangkok},
            postcode={10330},
            country={Thailand}}

\begin{abstract}
Recent studies in 3D object detection for autonomous vehicles aim to enrich features through the utilization of multi-modal setups or the extraction of local patterns within LiDAR point clouds.
However, multi-modal methods face significant challenges in feature alignment, and gaining features locally can be oversimplified for complex 3D object detection tasks.
In this paper, we propose a novel model, NV3D, which utilizes local features acquired from voxel neighbors, as normal vectors computed per voxel basis using \textit{K-nearest neighbors} (KNN) and \textit{principal component analysis} (PCA).
This informative feature enables NV3D to determine the relationship between the surface and pertinent target entities, including cars, pedestrians, or cyclists.
During the normal vector extraction process, NV3D offers two distinct sampling strategies: \textit{normal vector density-based sampling} and \textit{FOV-aware bin-based sampling}, allowing elimination of \textbf{up to 55\% of data while maintaining performance}.
In addition, we applied \textit{element-wise attention fusion}, which accepts voxel features as the query and value and normal vector features as the key, similar to the attention mechanism.
Our method is trained on the KITTI dataset and has demonstrated superior performance in car and cyclist detection owing to their spatial shapes.
In the validation set, NV3D without sampling achieves \textbf{86.60\%} and \textbf{80.18\%} mean Average Precision (mAP), greater than the baseline Voxel R-CNN by \textbf{2.61\%} and \textbf{4.23\%} mAP, respectively.
With both samplings, NV3D achieves \textbf{85.54\%} mAP in car detection, exceeding the baseline by \textbf{1.56\%} mAP, despite \textbf{roughly 55\% of voxels being filtered out}.

\end{abstract}

        


\end{frontmatter}



\section{Introduction}\label{sec1}


3D object detection is a task that allows computer machines to perceive real-world information through various sensors, including cameras~\cite {monoef,gs3d,caddn}, LiDAR~\cite{pointrcnn,voxel_rcnn,pv_rcnn}, radar~\cite{rv_fusion,rrpn}, etc.
Among these various sensors, cameras are frequently considered the most economically viable option, as they provide rich color features; however, their performance can be significantly impaired under unfavorable weather conditions.
While radar sensors can detect blind spots or occluded areas, their effectiveness can be limited due to their low resolution and poor localization, which may not meet requirements in certain applications.
LiDAR sensors can be regarded as the most viable option for advanced autonomous vehicles, as their intrinsic precise localization offers rich spatial features, which are highly reliable and advantageous for 3D object detection despite the lack of color features.
Recently, several methods~\cite{sfd,uvtr,virconv,graphalign2,is_fusion,objectfusion} have attempted to integrate LiDAR point clouds with camera RGB features, primarily to mitigate the issues of color feature absence in LiDAR information.
Nevertheless, a multiple data source setup causes issues with feature alignment, which continues to attract interest from research communities.

One of the major challenges for 3D object detection is designing efficient algorithms for machine perception of input data, such as point clouds and images, while balancing the trade-off between precision and speed.
Recent studies on 3D object detection for self-driving vehicles are broadly categorized into two main strategic approaches: single-modal detection and multi-modal detection.
In single-modal detection~\cite{f_pointnet,voxel_rcnn,faster_rcnn,monogrnet}, the system contains only one input, while a multi-modal system~\cite{avod,fconv,deepfusion,objectfusion} combines multiple inputs through the provision of feature fusion, thus improving the understanding of target objects and the surrounding environments.
The underlying idea of multi-modal detection is that each data source contains unique features, providing more comprehensive and reliable insights that potentially improve detection accuracy and robustness for real-world use~\cite{sfd,uvtr,virconv,graphalign2,is_fusion,objectfusion}.
However, as our study primarily focuses on analyzing point cloud features, for the sake of simplicity, we adhere to a single-modal setup to better facilitate our specific investigation of these properties.

Existing 3D detection methods based on LiDAR data can be broadly classified into two categories: point-based and voxel-based methods.
Point-based models, including PointNet++~\cite{pointnet2}, effectively process raw point clouds to capture fine-grained geometric details at a high computational cost.
PointRCNN~\cite{pointrcnn} extends this approach by using PointNet++~\cite{pointnet2} as its main backbone, while 3DSSD~\cite{3dssd} uses \textit{feature and distance farthest point sampling} with a \textit{candidate generation layer}, similar to VoteNet~\cite{votenet}.
On the other hand, voxel-based methods are inherently locally aware, as they partition points into a grid; however, the discretization potentially reduces positional accuracy and is restricted to a comprehensive local structural representation.
Early voxel-based approaches, such as SECOND~\cite{second} and Voxel R-CNN~\cite{voxel_rcnn}, introduced voxel-based backbone networks in conjunction with two-stage detection architectures.
Among recent methods, AFDetV2~\cite{afdetv2} proposes an anchor-free single-stage detection framework with a self-calibrated convolutional block and a spatial attention mechanism~\cite{sc_conv}.
Hybrid methods combining both point clouds and voxels, such as PV-RCNN~\cite{pv_rcnn}, integrate PointNet++\cite{pointnet2}-based point feature extraction with a voxel backbone based on SECOND\cite{second}.
Despite their considerable performance improvement, they generally face a high computational cost, which arises from neighbor searching and grouping, as well as the need for large architectures to handle points and their neighbors.
While the reduction of the neighbor searching and grouping costs remains a challenging trade-off, the large architectures can be addressed by feature engineering at an early stage.

To address this issue, this paper proposes NV3D, a low-complexity voxel-based 3D object detection model with two sampling techniques based on voxel feature density and normal vector density, thereby addressing ineffective neighbor features that lead to slower computation time.
The design of NV3D was based on two main hypotheses.
First, there is redundancy of dense point cloud clusters in the near distance.
Secondly, geometrical features from neighboring points can be leveraged to extract more relevant features.
Hence, NV3D utilizes normal vector features, which are extracted using the method proposed in~\cite{surf_recon}: applying \textit{K-nearest neighbors} (KNN) to determine neighboring points, followed by \ textit {principal component analysis} (PCA) to compute normal vectors.
This normal vector feature helps the model understand their surface orientation and direction at each corresponding point.

In addition, as discussed in VirConv~\cite{virconv}, distant points are more influential to model performance than nearby points for computing virtual points; this trend is also likely observed in real LiDAR point clouds.
Therefore, we propose \textit{FOV-aware bin-based sampling}, unlike \textit{general bin-based sampling} proposed in VirConv~\cite{virconv}, preserving the continuous density of point clouds, resulting in a smoother dropping strategy, and the \textit{normal vector density-based sampling}, which drops voxel features based on the majority of normal vector direction -- the largest flat plane or road for our environment.
To merge two information after two samplings, we adopt the attention mechanism~\cite{attn}: the key and value are voxel features, and the query is normal vector features, using element-wise attention fusion.
The main contribution of NV3D is to introduce the use of normal vector features extracted from local voxel features to sample relevant voxels, thereby enhancing the understanding of surface orientation for 3D object detection frameworks.

\section{Related Work}\label{sec2}

This section discusses related work regarding LiDAR-based 3D object detection, local feature-aware 3D object detection, and normal vector extraction approaches.

\subsection{LiDAR-based 3D Object Detection}

LiDAR point clouds are unordered 3D spatial features which can be constructed in two broad types: raw point clouds and voxel-based representation. 
The point-based representation is the early method of 3D object detection.
The influential work PointNet~\cite{pointnet}, requires max pooling as a symmetric function and returns features from the unstructured point clouds. 
PointRCNN~\cite{pointrcnn} adopts PointNet++~\cite{pointnet2} as its backbone, and achieve a two-stage 3D object detection framework.
Point-GNN~\cite{point_gnn} constructs a graph-based representation by connecting points within a fixed radius after voxel downsampling to reduce point cloud density.
3DSSD~\cite{3dssd} presents feature and distance farthest point sampling with a candidate generation layer, a method similar to VoteNet~\cite{votenet}.
SASA~\cite{sasa} introduces \textit{semantics-guided farthest point sampling} for enhanced feature mapping with input point coordinates.
LISO~\cite{liso} is a modern method that provides self-supervised learning to train object detection networks using unlabeled sequences of LiDAR point clouds which help reduce manual annotation.
However, the point-based approach introduces extra computational cost.

Another efficient approach is voxel-based representation, VoxelNet~\cite{voxelnet} introduces the end-to-end framework with the voxel-based approach: divide point clouds into grids each representing mean values of points, by randomly sampling point clouds from each voxel and performing \textit{voxel feature encoding} to obtain point-wise features. 
SECOND~\cite{second} presents a 3D backbone which is a core for modern voxel-based 3D object detection for autonomous vehicles.
Voxel R-CNN~\cite{voxel_rcnn} leverages \textit{voxel region of interest pooling} (ROI pooling) to extract RoI features from a 3D backbone based on 3D proposals, generated from applying BEV representation of the 3D backbone output features to a 2D backbone and \textit{region proposal network} (RPN). 
Therefore, RPN and RoI allow Voxel R-CNN~\cite{voxel_rcnn} to create a two-stage detection system.
AFDet~\cite{afdet} designs anchor-free one-stage detection, while its extended version, AFDetV2~\cite{afdetv2}, applies self-calibrated convolutional block and spatial attention mechanism~\cite{sc_conv}, as well as relocates \textit{intersection over union} (IoU) prediction head from 2D backbone to its anchor-free detector.
PDV~\cite{pdv} proposes density-aware ROI grid pooling through self-attention mechanism for 3D object detection.
Some recent studies, such as VirConv~\cite{virconv} and Re-VoxelDet~\cite{re_voxeldet}, propose architectural improvements for voxel-based 3D object detection.
VirConv~\cite{virconv} introduces a \textit{stochastic voxel discard} scheme to reduce computational load and a \textit{noise-resistant submanifold convolution} to mitigate noise from depth completion.
Re-VoxelDet~\cite{re_voxeldet} proposes a \textit{multi-view voxel backbone} to learn features from various perspectives, a \textit{hierarchical voxel-guided auxiliary neck} to help the model understand both semantic and spatial features, and a \textit{rotation-based group head} to separately classify vehicles and other objects.
Our work, NV3D, is also a voxel-based approach, adopting the backbone structure from Voxel R-CNN~\cite{voxel_rcnn}, which makes our model more efficient due to low-cost computation.
However, this method disregards the rich spatial features of point clouds, unlike PointNet++~\cite{pointnet2}.

\subsection{Local Feature-Aware 3D Object Detection}

Some contemporary 3D object detection approaches try to enhance detection performance by processing point clouds or voxels along with their neighbors. 
This technique helps the model recognize spatial shapes or neighbors, improving the overall performance.
PointRCNN~\cite{pointrcnn} builds upon PointNet++~\cite{pointnet2} to process point features along with their neighbors for enhanced 3D object detection.
VoteNet~\cite{votenet} introduces Hough voting from a deep network instead of KNN. 
The Hough voting of VoteNet~\cite{votenet} is computed from the base model of PointNet++~\cite{pointnet2} followed by shared-weight \textit{multilayer perceptron} (MLP) then KNN from voting and shared-weight PointNet~\cite{pointnet} to aggregate all voted features.
Point-GNN~\cite{point_gnn} connects surrounding points within a fixed radius, performs MLP, and aggregates with max pooling.
Voxel R-CNN~\cite{voxel_rcnn} also uses voxel query to group voxel features and voxel RoI pooling to enhance local feature awareness at the RoI layer.
For recent models, LoGoNet~\cite{logonet} improves 3D object detection by applying cross-modal local to global fusion, while EB+FCR~\cite{eb_fcr} attempts to improve the correlation based local feature point learning by correlation point embedding, both of which result in better feature representation in point clouds.
CIANet~\cite{cianet} applies \textit{channel-spatial hybrid attention} and \textit{contextual self-attention} module to generate higher feature 3D proposals enhancing a 2D backbone.
MSPV3D\cite{mspv3d} fuses multi-scale voxel features of different sizes within its backbone and utilizes the detection head from PV-RCNN~\cite{pv_rcnn} for final 3D object detection.
Although these modern methods focus on directly processing neighbor features, these features can be transformed into more informative representation to fully exploit their potential.
NV3D recognizes this gap and proposes normal vectors as key features for local-aware 3D object detection.
Our method can reduce memory consumption by avoiding storing all local points or voxels but transforming them into normal vectors.
This also helps the model become aware of surface orientation through the direction of normal vectors.

\subsection{Normal Vector Extraction Approaches}

There are two primary approaches for estimating this feature: classical methods and deep learning-based techniques.
Local PCA~\cite{surf_recon} is a simple way to extract the normal vector of each point cloud by gathering \textit{k} nearest points and computing the normal vector via PCA.
RANSAC~\cite{ransac} is an iterative method for determining the best-fit plane, where the vector perpendicular to the plane represents the normal vector.
Also, RANSAC~\cite{ransac} is effective in dealing with outliers.

Modern methods attempt to get deep learning involved by processing real-world coordinates of point clouds.
PCPNet~\cite{pcpnet} is a pioneering method in this field.
By using multi-scale and spatial transformers, PCPNet~\cite{pcpnet} can achieve invariant to point ordering for normal vectors estimation.
Nesti-Net~\cite{nesti_net} proposed an eminent framework by transforming local neighborhoods into structured representations and applying a convolutional neural network.
DeepFit~\cite{deepfit} leverages point-wise weights for weighted least squares surface fitting, which are extracted using PointNet~\cite{pointnet} and its global feature output.
SHSNet~\cite{shsnet} introduces \textit{patch encoding} and \textit{shape encoding} serving as the generation of normal orientation signs.

To some extent, the time consumption of traditional and deep learning-based methods can be comparable depending on the hardware and algorithm.
However, the performance of deep learning-based methods can be more accurate, since modern architectures allow us to deal with noised data and understand real-world spatial features, unlike traditional methods.
Nonetheless, NV3D is strict to the classical method by performing KNN followed with PCA and selecting the least variance vector as a normal vector, as achieved in~\cite{surf_recon}.

\section{Methodology}

NV3D involves a state-of-the-art feature extraction method for 3D object detection using normal vectors extracted from local features. 
The key components of NV3D include the normal vector extraction module, normal vector density-based sampling, FOV-aware bin-based sampling, and element-wise attention fusion.
NV3D adopts a backbone and detection head from Voxel R-CNN~\cite{voxel_rcnn}.
However, the architecture is enhanced by using normal vector features, including normal vector directions and their density in normal vector space.
The inputs to NV3D are a set of voxel features \(\{ \mathbf{v}_i = (x_i, y_i, z_i, r_i) \}_{i=1}^N\) extracted by a mean of point clouds inside each voxel and a set of normal vectors from local voxels \(\{ n_i = (a_{x_i}, a_{y_i}, a_{z_i}, d_i) \}_{i=1}^N\).

\begin{figure*}[!ht]
        \centering
        \includegraphics*[width=\linewidth]{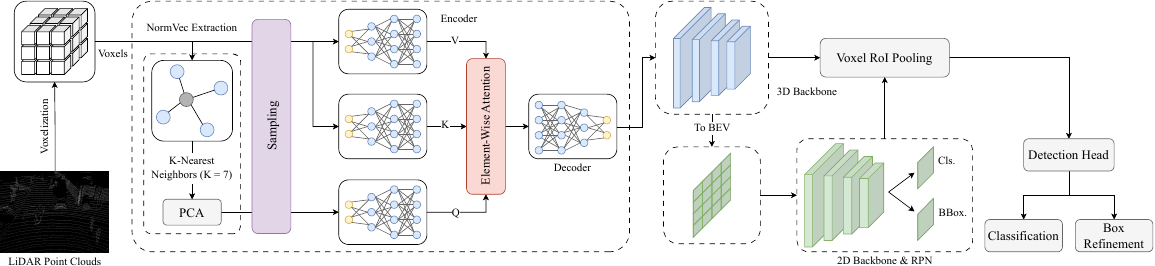}
        \caption{
                The architecture of NV3D for 3D object detection. 
                The LiDAR point clouds are first divided into voxels. 
                These voxel features are used to compute normal vector features, as well as sampling masks through the normal vector extraction module.
                Sampling masks are applied before feeding voxel and normal features into element-wise attention fusion to merge two features.
                The merged feature is continued in Voxel R-CNN-based architecture for a complete 3D object detection model.
        }
        \label{fig:overview_arch}
\end{figure*}

As shown in Fig.~\ref{fig:overview_arch}, NV3D first converts points into voxels, and extracts normal vectors by the normal vector extraction module; during which process, normal vector properties allow sampling voxels before being fed into the backbone. 
In addition, the sampled voxel and normal vector features are processed in sequence of MLP and element-wised attention mechanism to provide the data-relevant score for further detection.

\subsection{Normal Vector Extraction}

\begin{figure}[!ht]
        \centering
        \begin{subfigure}[b]{0.47\linewidth}
            \centering
            \includegraphics[width=\linewidth]{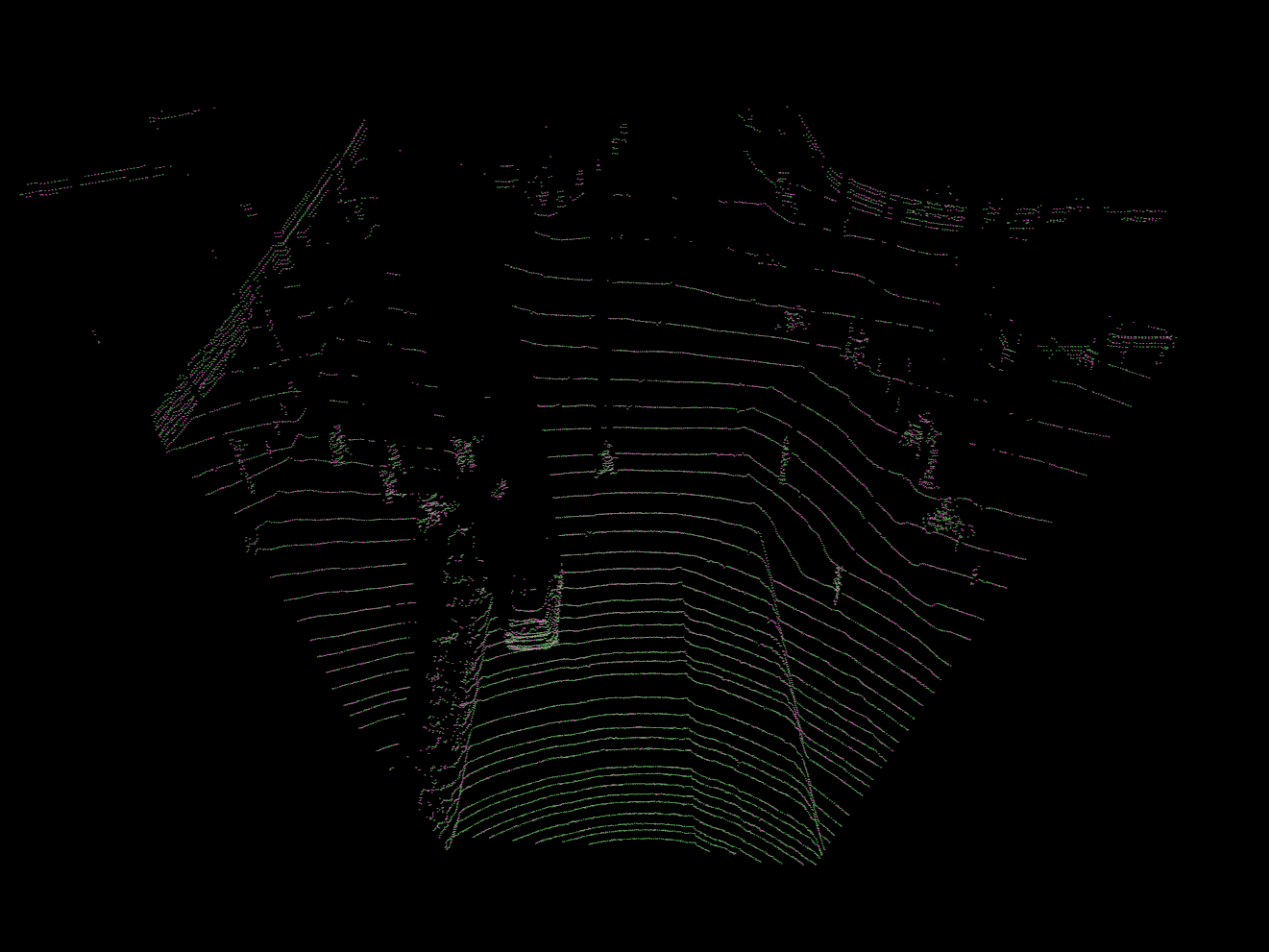}
            \caption{}
        \end{subfigure}
        \hfill
        \begin{subfigure}[b]{0.47\linewidth}
            \centering
            \includegraphics[width=\linewidth]{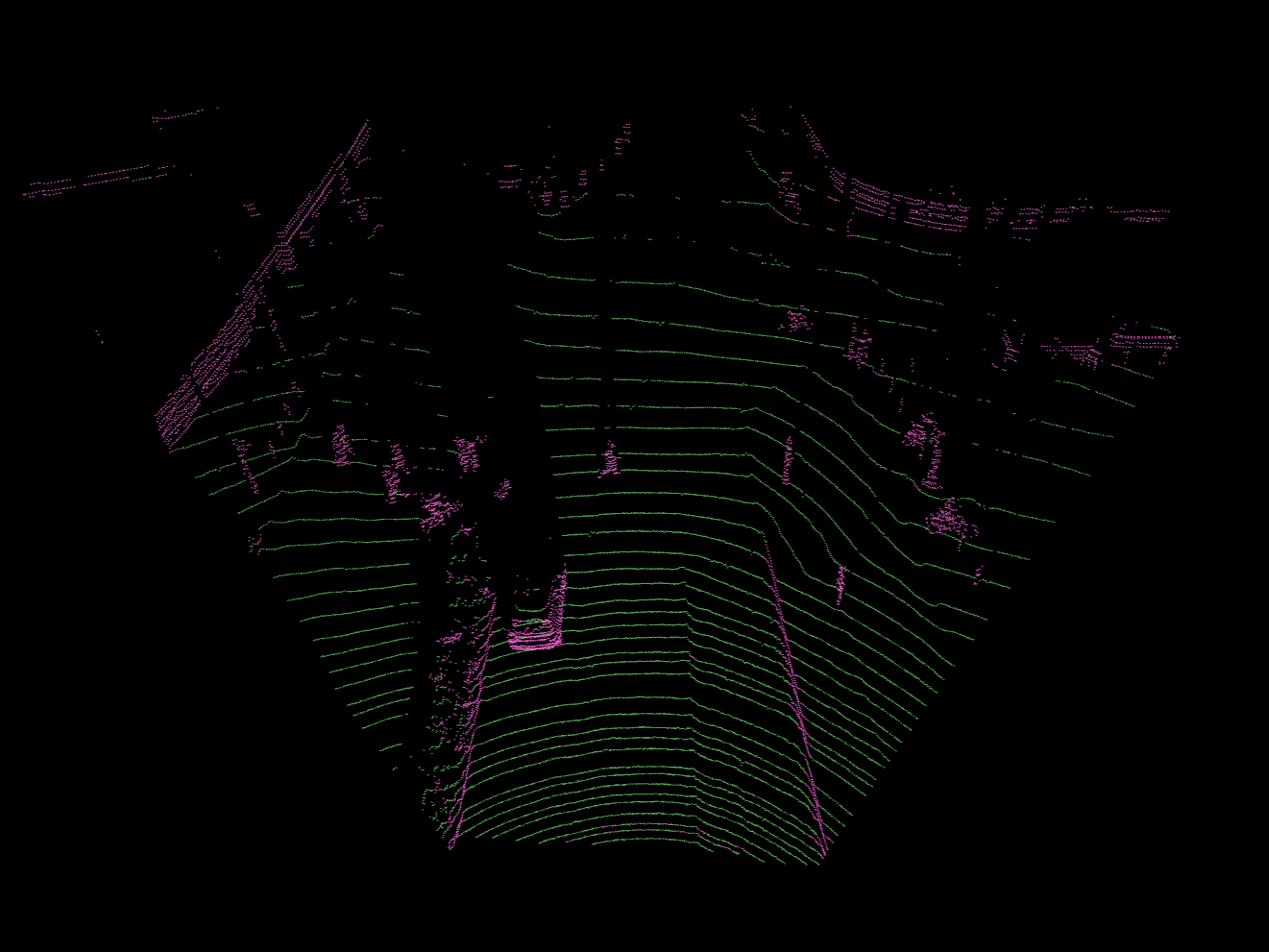}
            \caption{}
        \end{subfigure}
        \caption{
                (a) Normal vectors directly extracted from points inside each voxel.
                (b) Normal vectors extracted from surrounding voxels.
        }
        \label{fig:voxel_to_vector}
\end{figure}

Given that each voxel typically contains multiple points, computing a unit normal vector for each voxel is feasible.
However, this straightforward approach comes with two drawbacks that restrict it from generating reliable estimates of unit normal vectors:
1. voxels that contain less than three point clouds and 2. the alignment of point clouds inside a voxel. 
There are less than 20\% of voxels containing more than three points.
Even though using pseudo point clouds and ignoring voxels that contain less than three points, the second problem still arises.
As illustrated in Fig.~\ref{fig:voxel_to_vector}, the consistency of normal vector extraction exhibits variation. The color of each point in the figure represents the z-axis value of the normal vector, ranging from low to high (magenta to green).

NV3D adopts Local PCA~\cite{surf_recon} to solve both problems.
Neighbor points \(\{ \mathbf{nb}_i = \{(x_{ij}, y_{ij}, z_{ij})\}_{j=1}^k \}_{i=1}^N\) are extracted from KNN.
For this work, we use \(K=7\) to retrieve neighbors, \(\{ \mathbf{nb}_i = \{(x_{ij}, y_{ij}, z_{ij})\}_{j=1}^7 \}_{i=1}^N\).
The normal vector computation of each point is achieved by performing PCA. 
However, we use only the axis with the lowest variance, whose corresponding eigenvector represents the point's normal vector.
To obtain the density of normal vectors, we define a spherical volume with a fixed radius of 0.25 unit length and count the normal vectors in normal vector space inside each sphere, see Fig.~\ref{fig:norm_drop_explanation}, then normalize it for each batch.
As a result, normal vector features \( \{ \mathbf{n}_i = (nx_i, ny_i, nz_i, d_i) \}_{i=1}^N\) are computed for each batch, and used as one of the inputs to NV3D.

\subsection{Input Sampling Method}

Sampling is the main key to reducing data size leading to faster computation. 
Two fundamental methods are 1. random sampling and 2. farthest point sampling.
As discussed in VirConv~\cite{virconv}, both methods remove points across all ranges, including distant ones.
However, distant point clouds contribute significantly to detection accuracy, whereas nearby virtual point clouds have less impact.
Hence, both random and farthest sampling may not be suitable for 3D object detection task, and VirConv~\cite{virconv} proposed bin-based sampling to tackle the problem. 
In our work, although data source used in VirConv~\cite{virconv} differs from LiDAR point clouds, a similar trend may be expected.

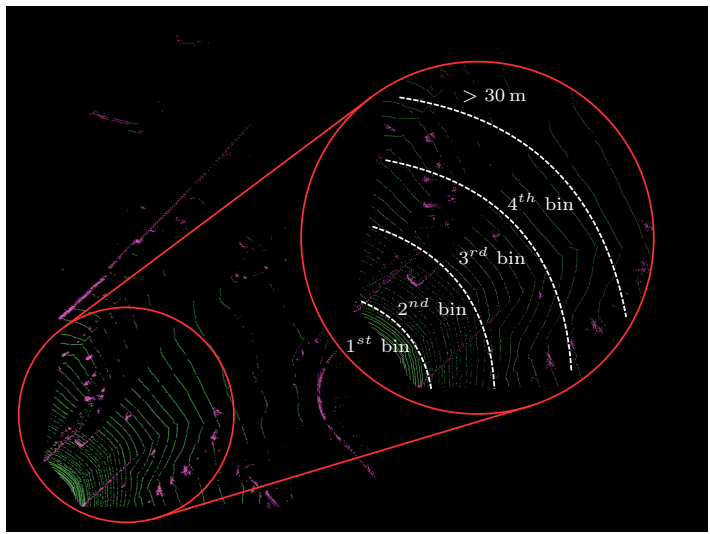
\begin{figure*}[!ht]
        \centering
        \begin{overpic}[width=0.68\linewidth]{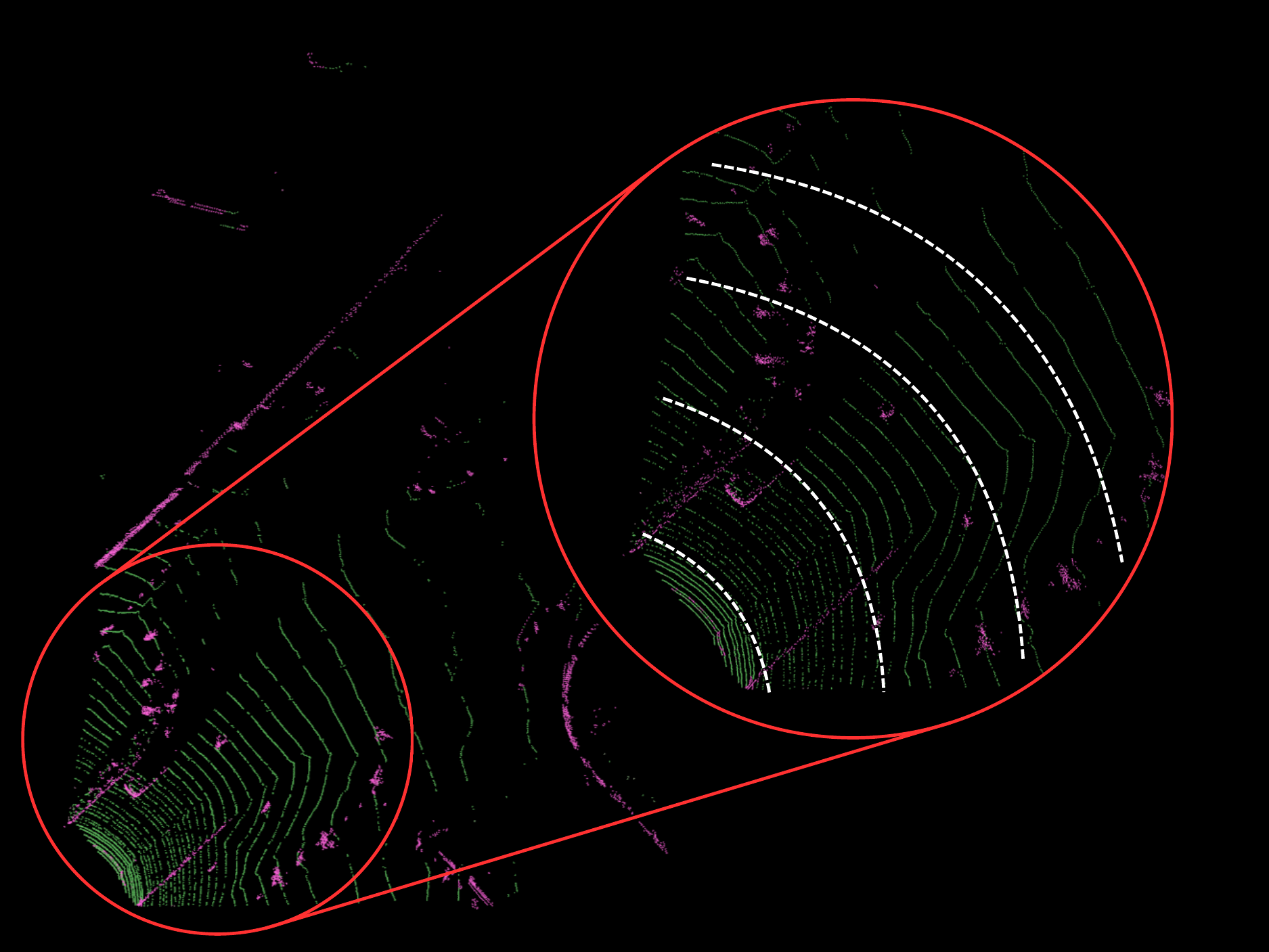}
                \put(65,61.5){\fontsize{7pt}{12pt}\selectfont\textcolor{white}{\(> 30 \, \mathrm{m}\)}}  
                \put(71.5,46){\fontsize{7pt}{12pt}\selectfont\textcolor{white}{\(4^{th}\) bin}}  
                \put(64.5,38.3){\fontsize{7pt}{12pt}\selectfont\textcolor{white}{\(3^{rd}\) bin}}  
                \put(55.8,31){\fontsize{7pt}{12pt}\selectfont\textcolor{white}{\(2^{nd}\) bin}}  
                \put(48.4,25.7){\fontsize{7pt}{12pt}\selectfont\textcolor{white}{\(1^{st}\) bin}}  
        \end{overpic}
        \caption{Visualization of normal vector density-based sampling followed by FOV-aware bin-based sampling showing the continuous density of voxel feature.}
        \label{fig:general_bin_based}
    \end{figure*}

\begin{figure}[!ht]
        \centering
        \begin{subfigure}[b]{.32\linewidth}
                \centering
                \includegraphics[width=\linewidth]{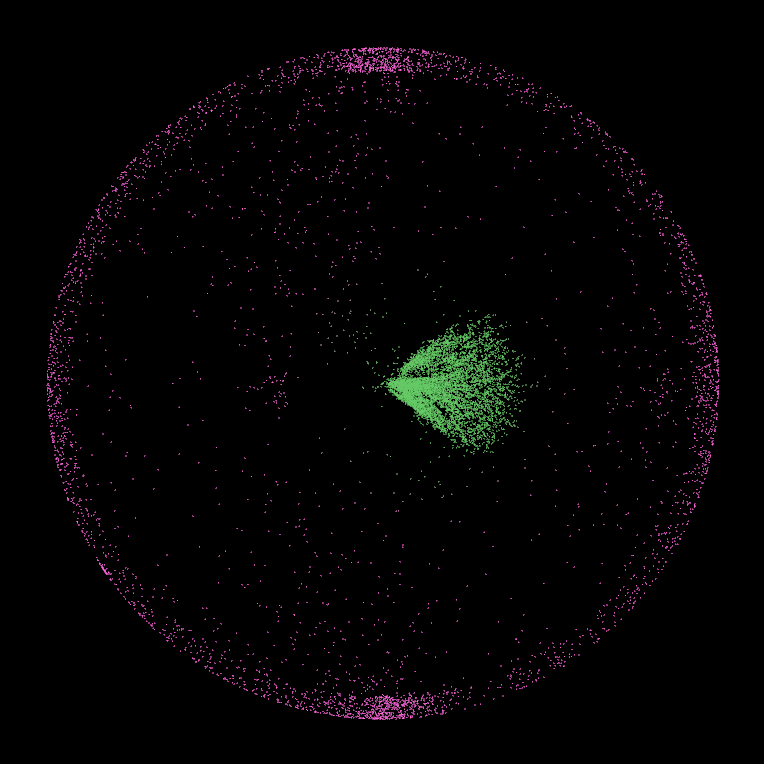}
                \caption{}
        \end{subfigure}
        \hfill
        \begin{subfigure}[b]{.32\linewidth}
                \centering
                \includegraphics[width=\linewidth]{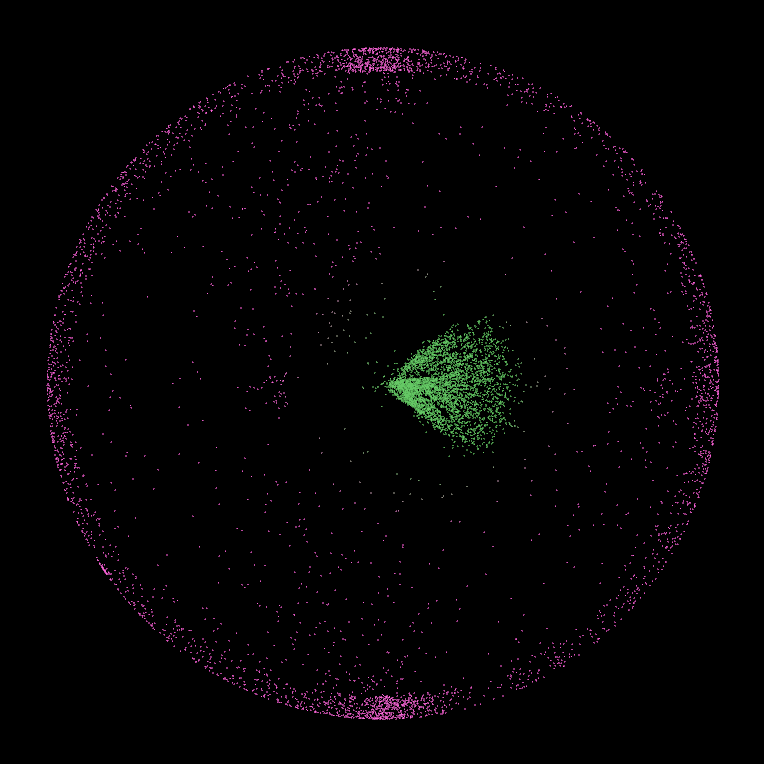}
                \caption{}
        \end{subfigure}
        \hfill
        \begin{subfigure}[b]{.32\linewidth}
                \centering
                \includegraphics[width=\linewidth]{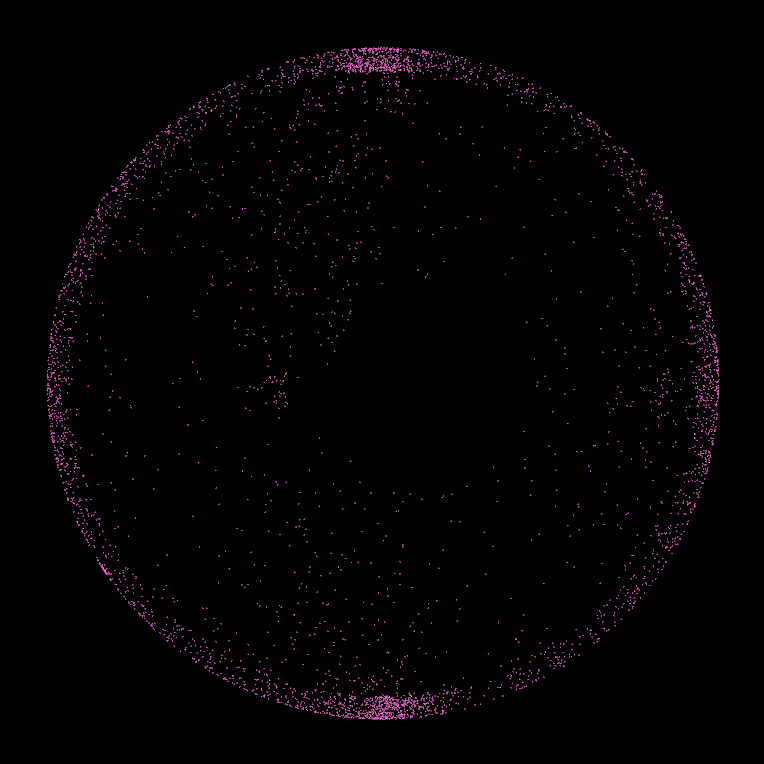}
                \caption{}
        \end{subfigure}
        \caption{Visualization of normal vectors for the same frame with varying drop rates, considering only normalized normal vector densities greater than 0.7: (a) original plot (no drop) (b) 50\% drop rate (c) 100\% drop rate.}
        \label{fig:norm_drop_explanation}
\end{figure}

With the bin-based sampling, the number of voxels per bin can be dropped to a specific amount, thereby eliminating redundant voxels.
Nonetheless, the general bin-based sampling creates non-continuous voxel feature density since it lacks fov awareness, as illustrated in Fig.~\ref{fig:general_bin_based}.
Hence, NV3D solves this issue with two sampling methods: 1. normal vector density-based sampling and 2. FOV-aware bin-based sampling.

\textbf{Normal vector density-based sampling} To sample points efficiently, we need to eliminate most of the redundant points -- road plane for our environment.
Most road planes are undeniably flat and consistent, this property leads to the normal vector density-based sampling method. 
The normal vector of each voxel is plotted on a unit sphere, as shown in Fig. 4, with colors representing the normalized density of normal vectors, ranging from low to high density (magenta to green).
In our case, only 50\% of the normal vectors with a density greater than 0.7 are dropped.
By setting a high threshold and removing only 50\%, we can prevent the loss of important feature points, such as those corresponding to cars.

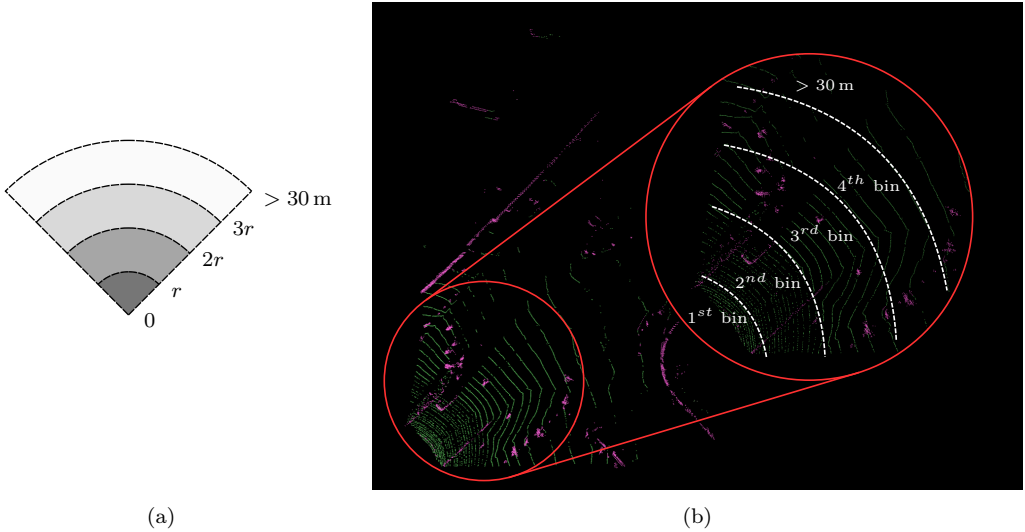
\begin{figure}[!ht]
        \centering
        \begin{subfigure}[b]{.33\linewidth}
                \centering
                \vspace*{0.355\linewidth}
                \begin{overpic}[width=\linewidth]{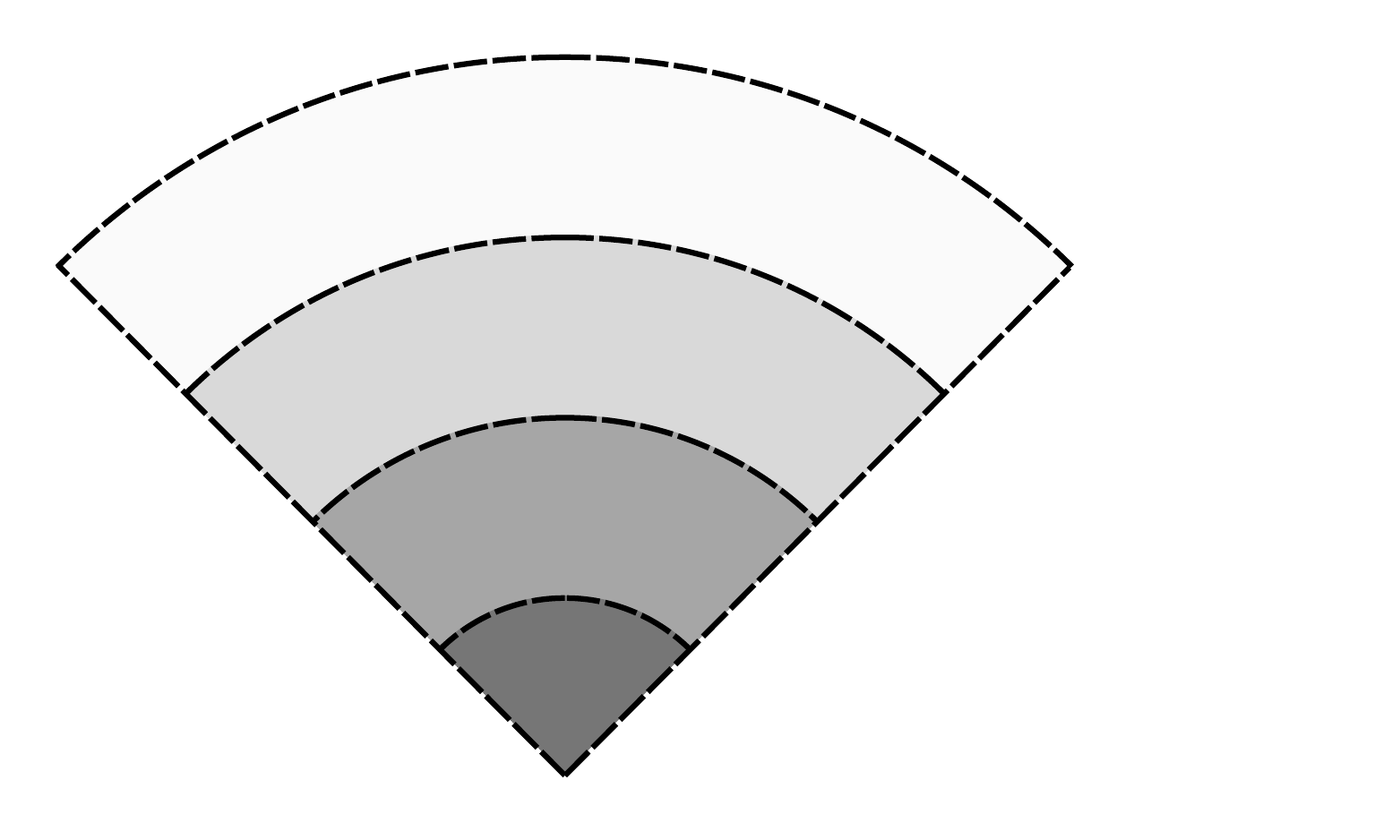}
                        \put(45, 1){\fontsize{8pt}{12pt}\selectfont\textcolor{black}{0}}  
                        \put(53, 10){\fontsize{8pt}{12pt}\selectfont\textcolor{black}{\(r\)}}  
                        \put(62, 19){\fontsize{8pt}{12pt}\selectfont\textcolor{black}{\(2r\)}}  
                        \put(71, 28){\fontsize{8pt}{12pt}\selectfont\textcolor{black}{\(3r\)}}
                        \put(80, 37){\fontsize{8pt}{12pt}\selectfont\textcolor{black}{\(> 30 \, \mathrm{m}\)}}  
                \end{overpic}
                \vspace*{0.355\linewidth}
                \caption{}
                \label{fig:fov_aware_explanation}
        \end{subfigure}
        \hfill
        \begin{subfigure}[b]{0.63\textwidth}
                \centering
                \begin{overpic}[width=\linewidth]{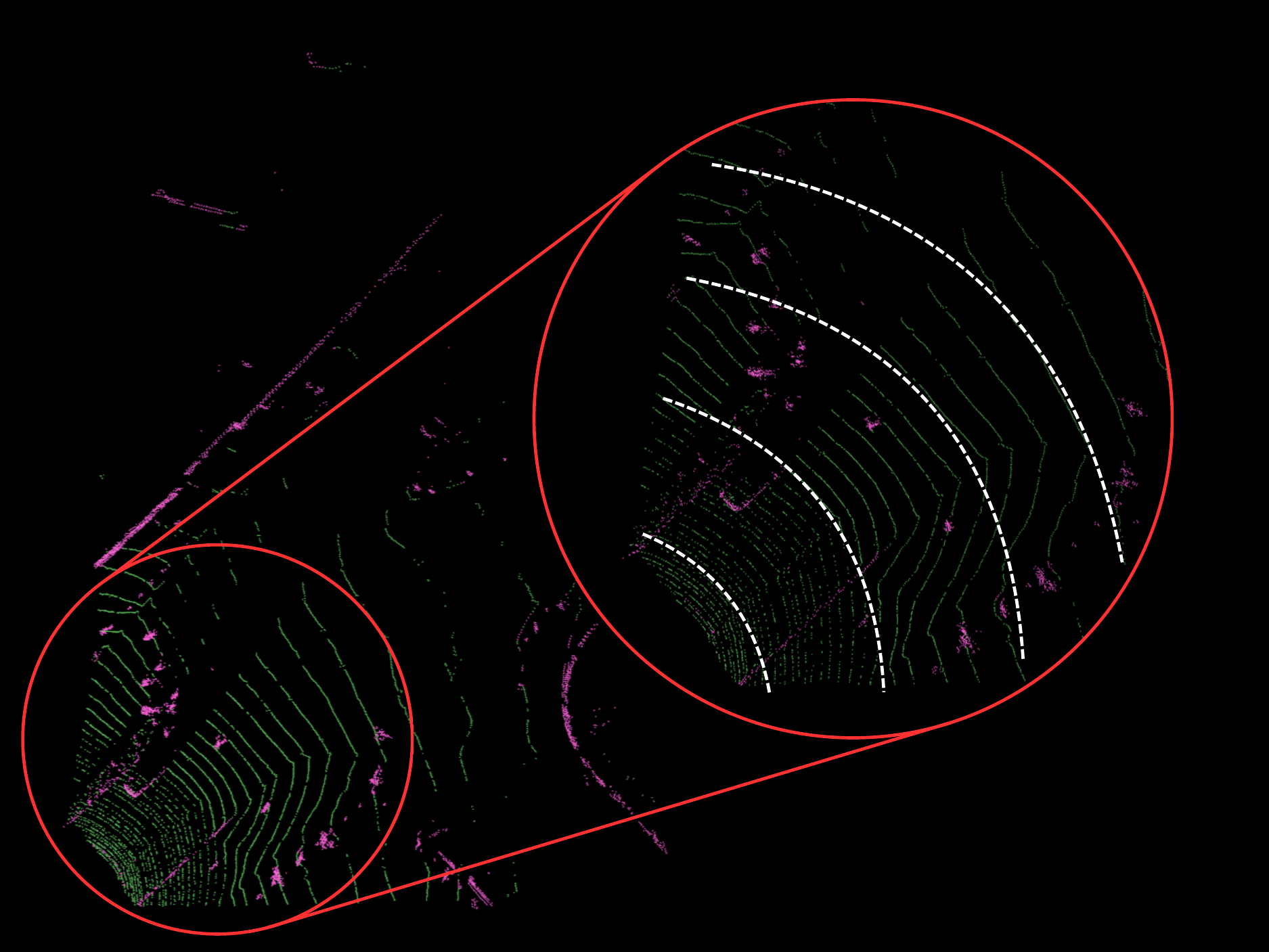}
                        \put(65,61.5){\fontsize{6.5pt}{12pt}\selectfont\textcolor{white}{\(> 30 \, \mathrm{m}\)}}  
                        \put(71.5,46){\fontsize{6.5pt}{12pt}\selectfont\textcolor{white}{\(4^{th}\) bin}}  
                        \put(64.3,38.3){\fontsize{6.5pt}{12pt}\selectfont\textcolor{white}{\(3^{rd}\) bin}}  
                        \put(55.8,31){\fontsize{6.5pt}{12pt}\selectfont\textcolor{white}{\(2^{nd}\) bin}}  
                        \put(48.4,25.7){\fontsize{6.5pt}{12pt}\selectfont\textcolor{white}{\(1^{st}\) bin}}  
                \end{overpic}
                \caption{}
            \end{subfigure}

        \caption{(a) FOV-aware sampling: \(\text{n}^\text{th}\) bin contains \(500 \times (2n - 1)\) points. (b) Visualization of FOV-aware bin-based drop showing the more consistency of voxel feature density.}
        \label{fig:fov_aware_sampling}
\end{figure}

\begin{figure*}[!ht]
        \centering
        \begin{subfigure}[b]{0.24\linewidth}
                \centering
                \includegraphics[width=\linewidth]{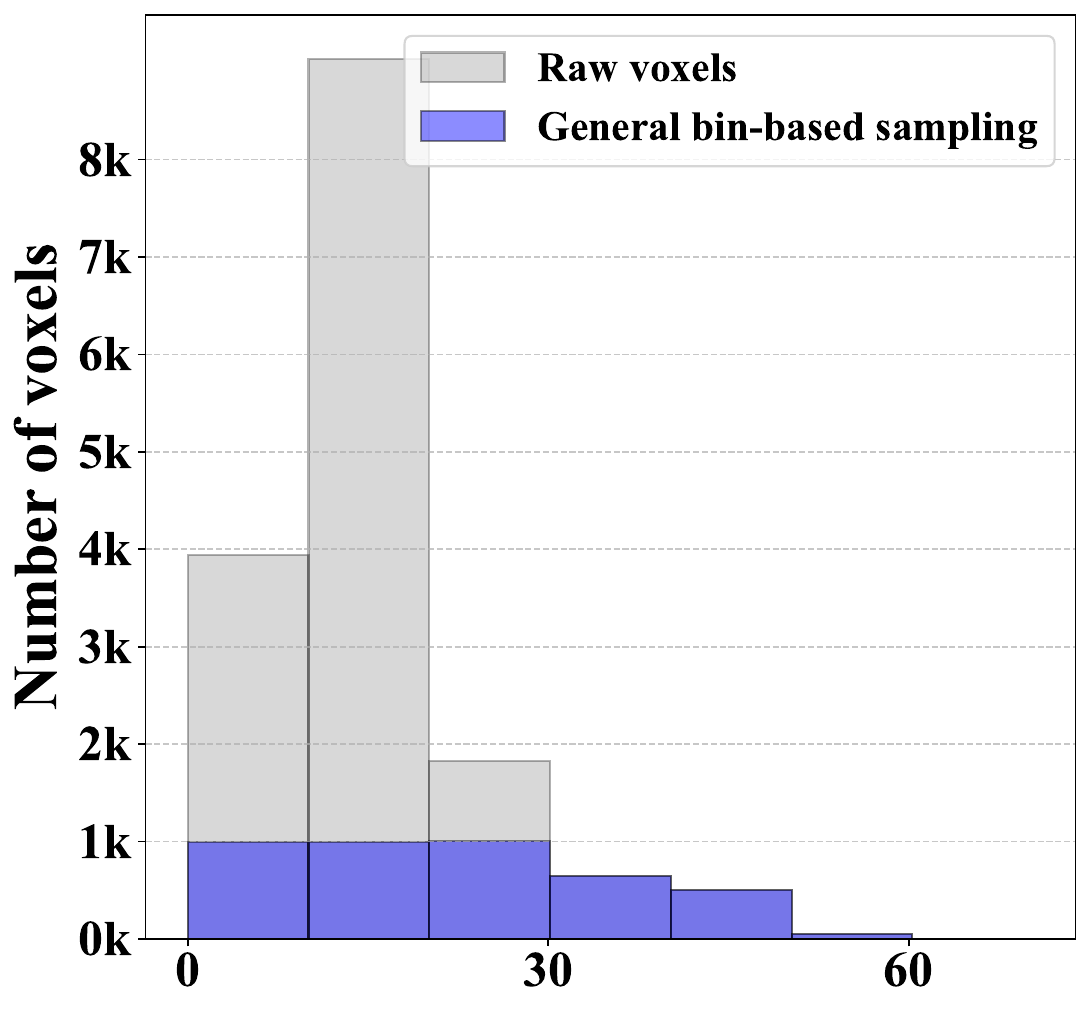}
                \caption{}
        \end{subfigure}
        \hfill
        \begin{subfigure}[b]{0.24\linewidth}
                \centering
                \includegraphics[width=\linewidth]{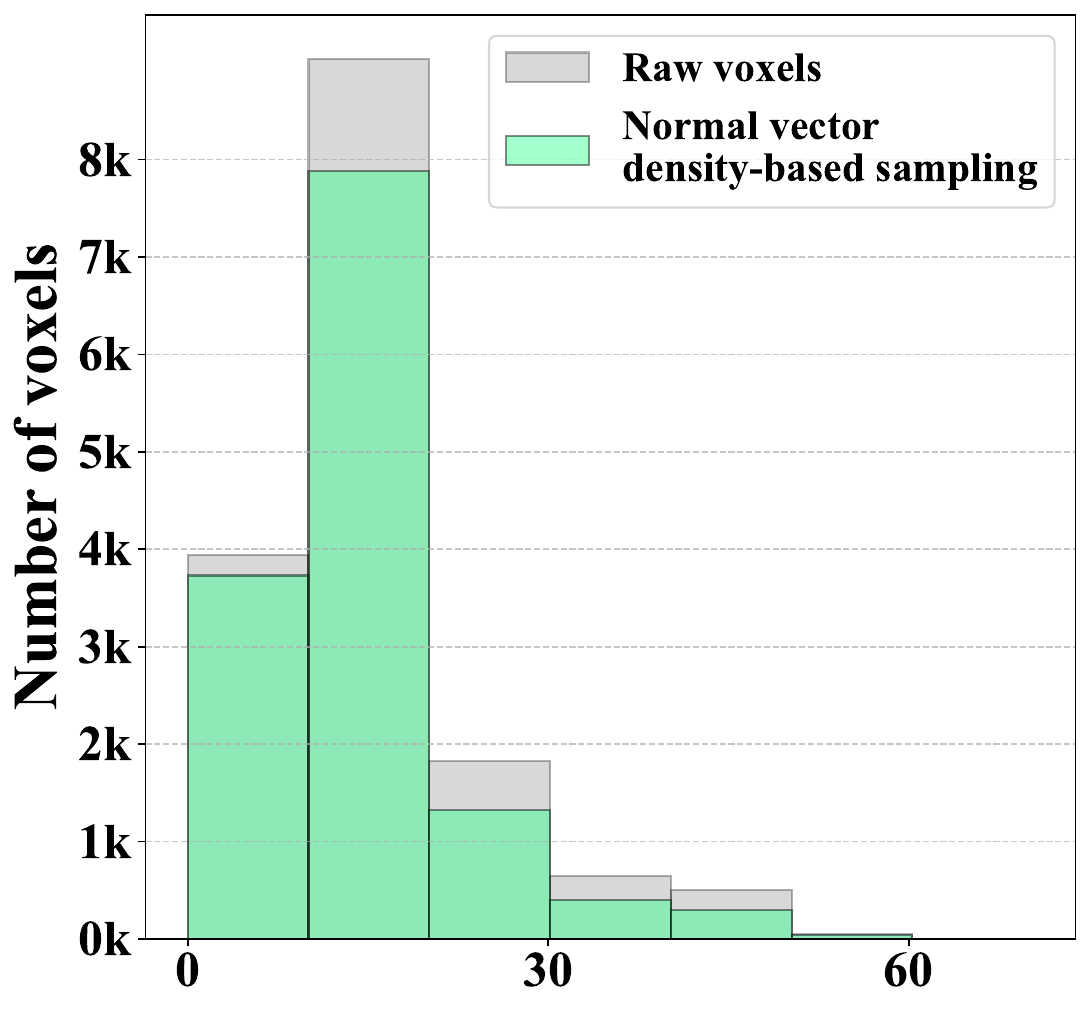}
                \caption{}
        \end{subfigure}
        \hfill
        \begin{subfigure}[b]{0.24\linewidth}
                \centering
                \includegraphics[width=\linewidth]{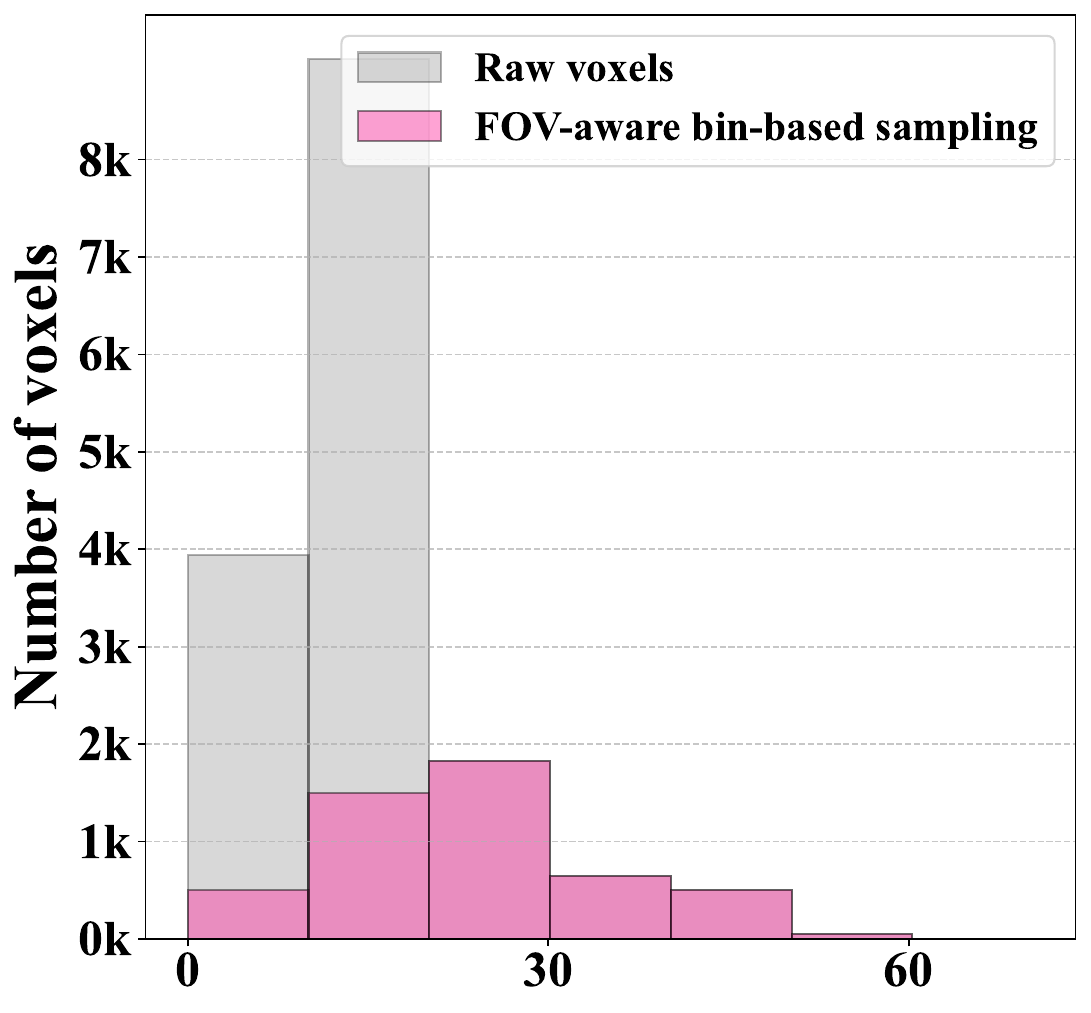}
                \caption{}
        \end{subfigure}
        \hfill
        \begin{subfigure}[b]{0.24\linewidth}
                \centering
                \includegraphics[width=\linewidth]{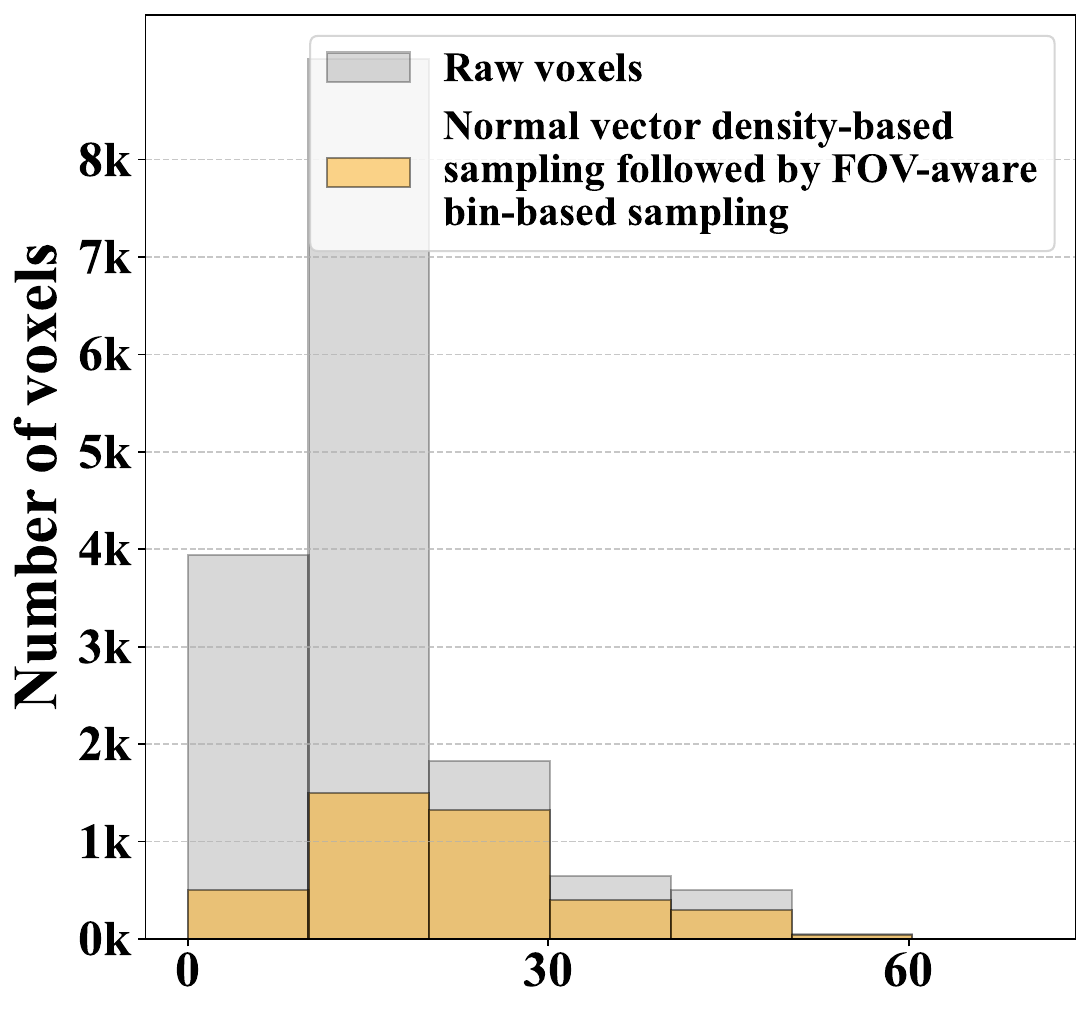}
                \caption{}
        \end{subfigure}
        \caption{Comparison of sampling methods presented through histograms, where the horizontal axis represents the distance from the LiDAR sensor, and the vertical axis indicates the number of voxels.}
        \label{fig:sampling_methods_pc}
\end{figure*}

\textbf{FOV-aware bin-based Sampling} As discussed in VirConv~\cite{virconv}, nearby virtual point clouds less contribute to the overall accuracy while distant point clouds much contribution. 
As seen in Fig.~\ref{fig:sampling_methods_pc}, around 90\% of voxels are condensed in the first 30 m, leaving the rest of 10\% of voxels for 30 m to out of rage (\(\approx\) 60 m).
VirConv~\cite{virconv} proposed a bin-based droppring strategy for nearby points while maintaining all distant points.
However, this unsophisticated bin-based method, despite its high speed and awareness of the loss in distant points.
Unlike random sampling and farthest sampling, it disrupts spatial continuity and creates a non-smooth transition between each bin, seen in Fig.~\ref{fig:general_bin_based}.
In contrast, the FOV-aware bin-based sampling is a fast and voxel feature density-aware dropping method that aims to ensure that nearby points have the same number of voxel features per bin area.
In Fig.~\ref{fig:fov_aware_explanation}, we assign the first bin to contain 500 points, resulting in a voxel feature density of \(\frac{500}{\pi r^2}\).
To maintain the same voxel feature density at \(n^{th}\) bin, we calculate the required number of voxel features as: \(\frac{500}{\pi * r^2} \times (\pi (nr)^2 - \pi (n(r-1))^2) = 500 \times (2n - 1) \).
Our sampling uses 10 bins, and we assign \(500 \times (2n - 1)\) points to the \(n^{th}\) bin accordingly.

\begin{figure*}[!ht]
        \centering
        \begin{subfigure}[b]{0.24\linewidth}
            \centering
            \includegraphics[width=\linewidth]{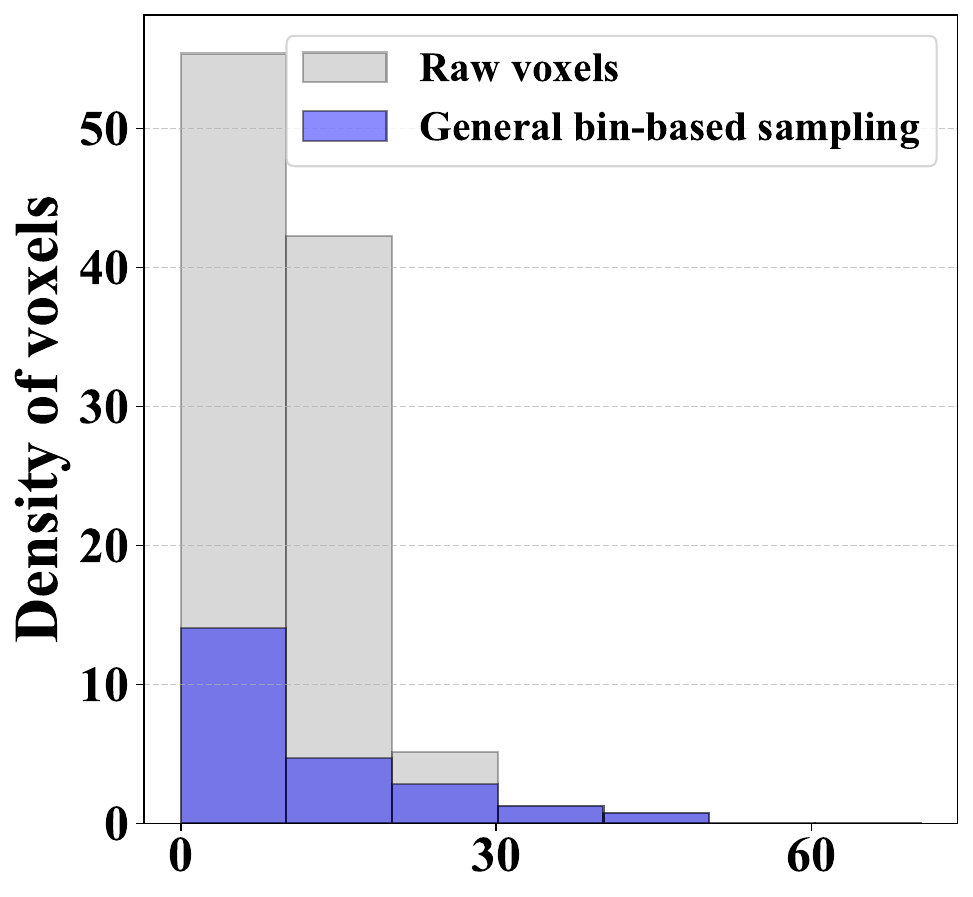}
            \caption{}
        \end{subfigure}
        \hfill
        \begin{subfigure}[b]{0.24\linewidth}
            \centering
            \includegraphics[width=\linewidth]{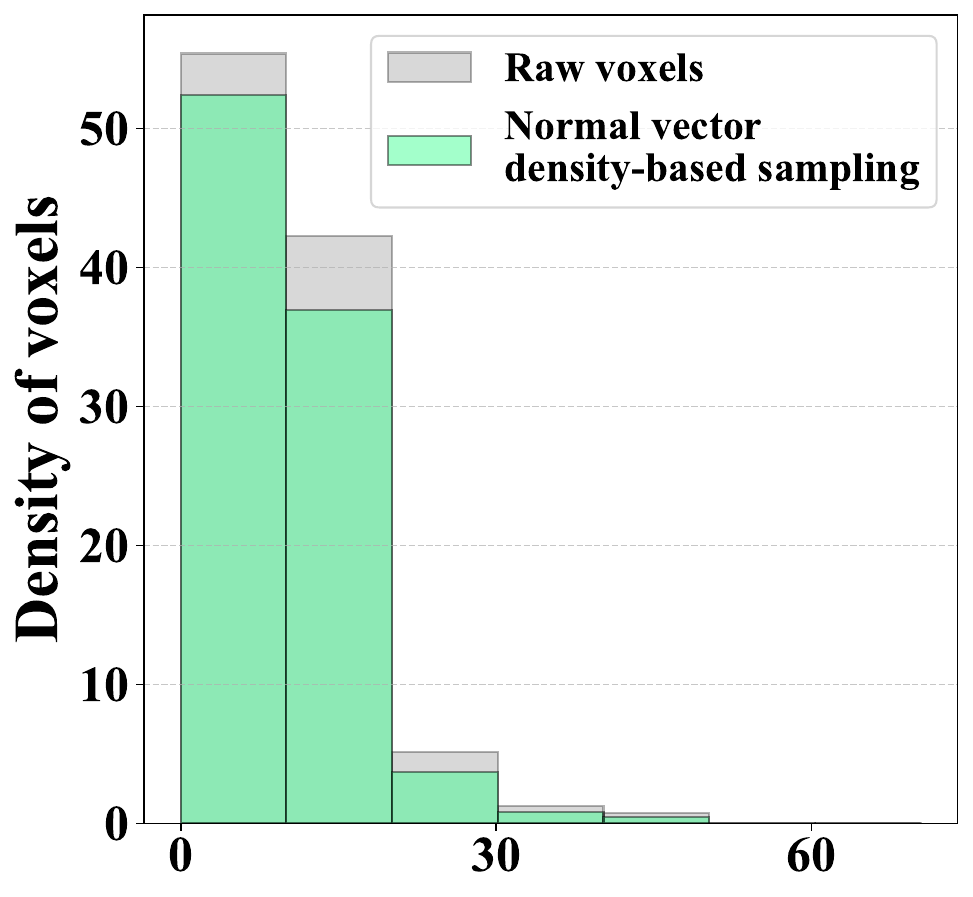}
            \caption{}
        \end{subfigure}
        \begin{subfigure}[b]{0.24\linewidth}
            \centering
            \includegraphics[width=\linewidth]{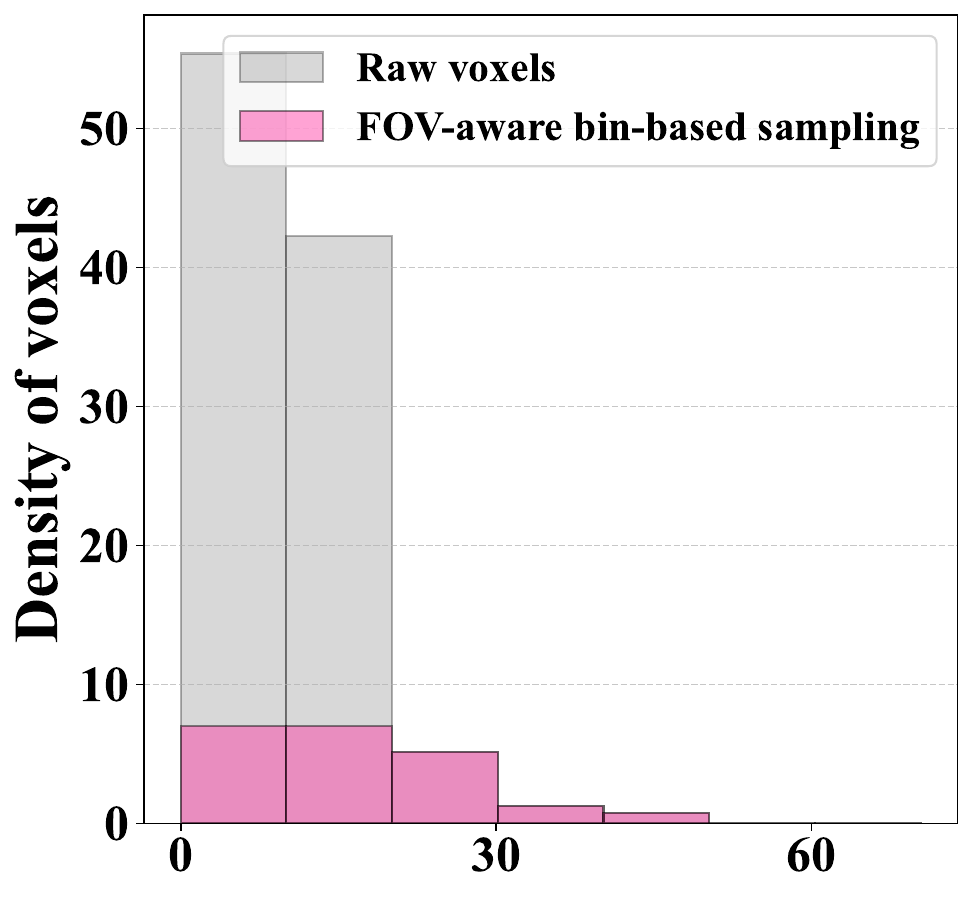}
            \caption{}
        \end{subfigure}
        \hfill
        \begin{subfigure}[b]{0.24\linewidth}
            \centering
            \includegraphics[width=\linewidth]{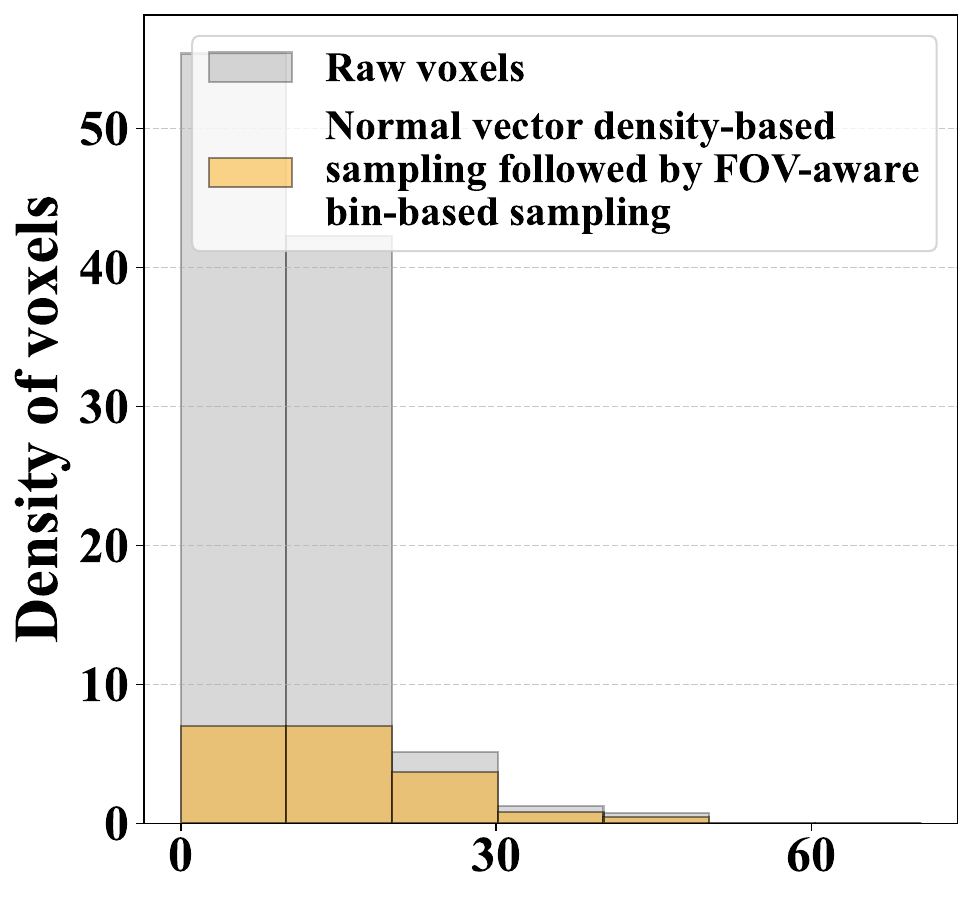}
            \caption{}
        \end{subfigure}
        \caption{Comparison of sampling methods presented through histograms, where the horizontal axis represents the distance from the LiDAR sensor, and the vertical axis indicates the density of voxel feature}
        \label{fig:sampling_methods_den}
\end{figure*}

Both normal vector density-based and FOV-aware bin-based sampling offer continuous sampling strategies for real-world data points.
Also, normal vector density-based sampling helps to remove redundant information from voxel features that contain similar normal vector directions, which mostly correspond to flat road surfaces.
The graphical explanation can be found in Fig.~\ref{fig:sampling_methods_pc} and Fig.~\ref{fig:sampling_methods_den} showing the number of voxels and the density of voxel feature for each bin, respectively.

\begin{figure*}[!ht]
        \centering
        \begin{subfigure}[b]{0.48\textwidth}
                \centering
                \includegraphics[width=\linewidth]{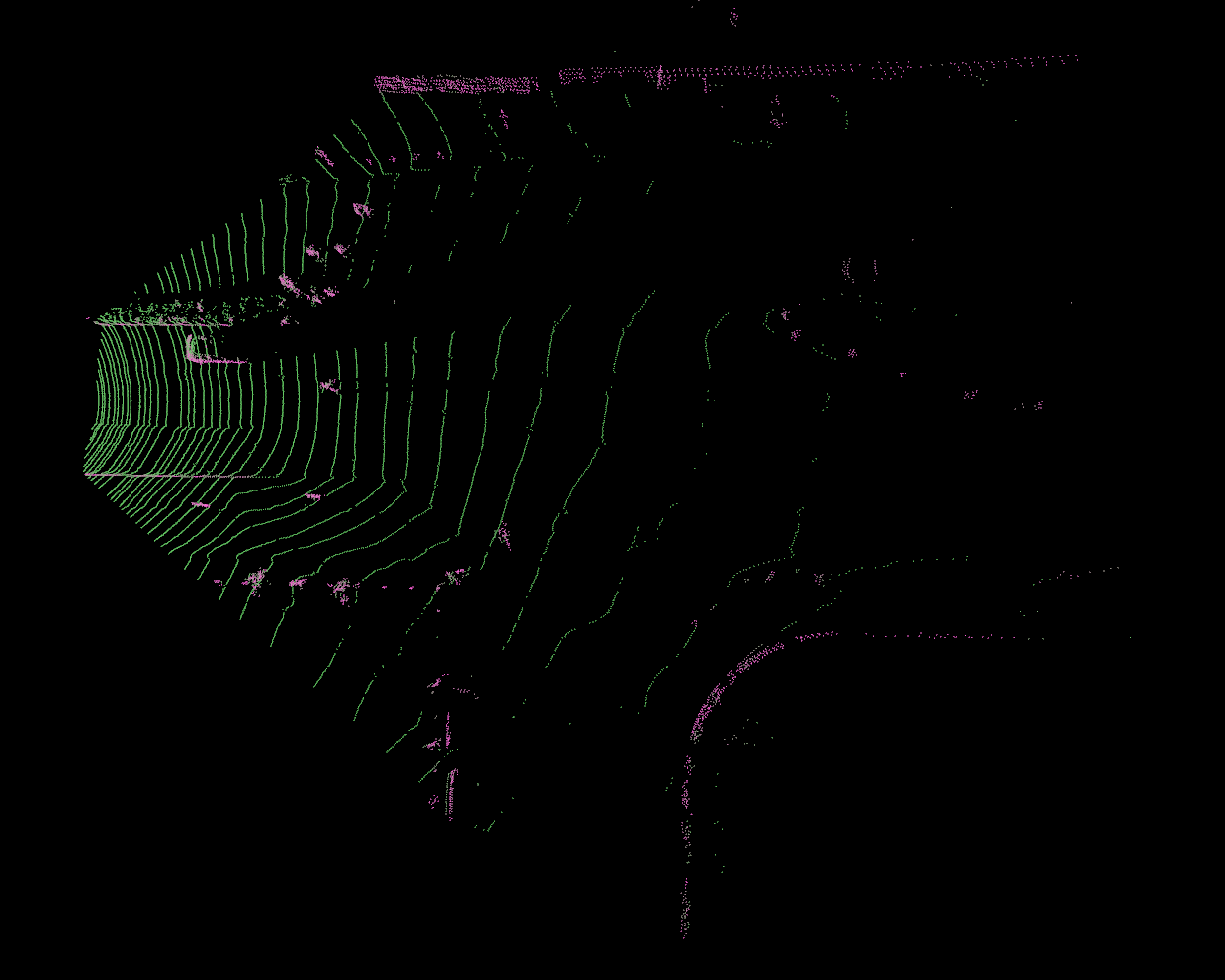}
                \caption{}
        \end{subfigure}
        \hfill
        \begin{subfigure}[b]{0.48\textwidth}
                \centering
                \includegraphics[width=\linewidth]{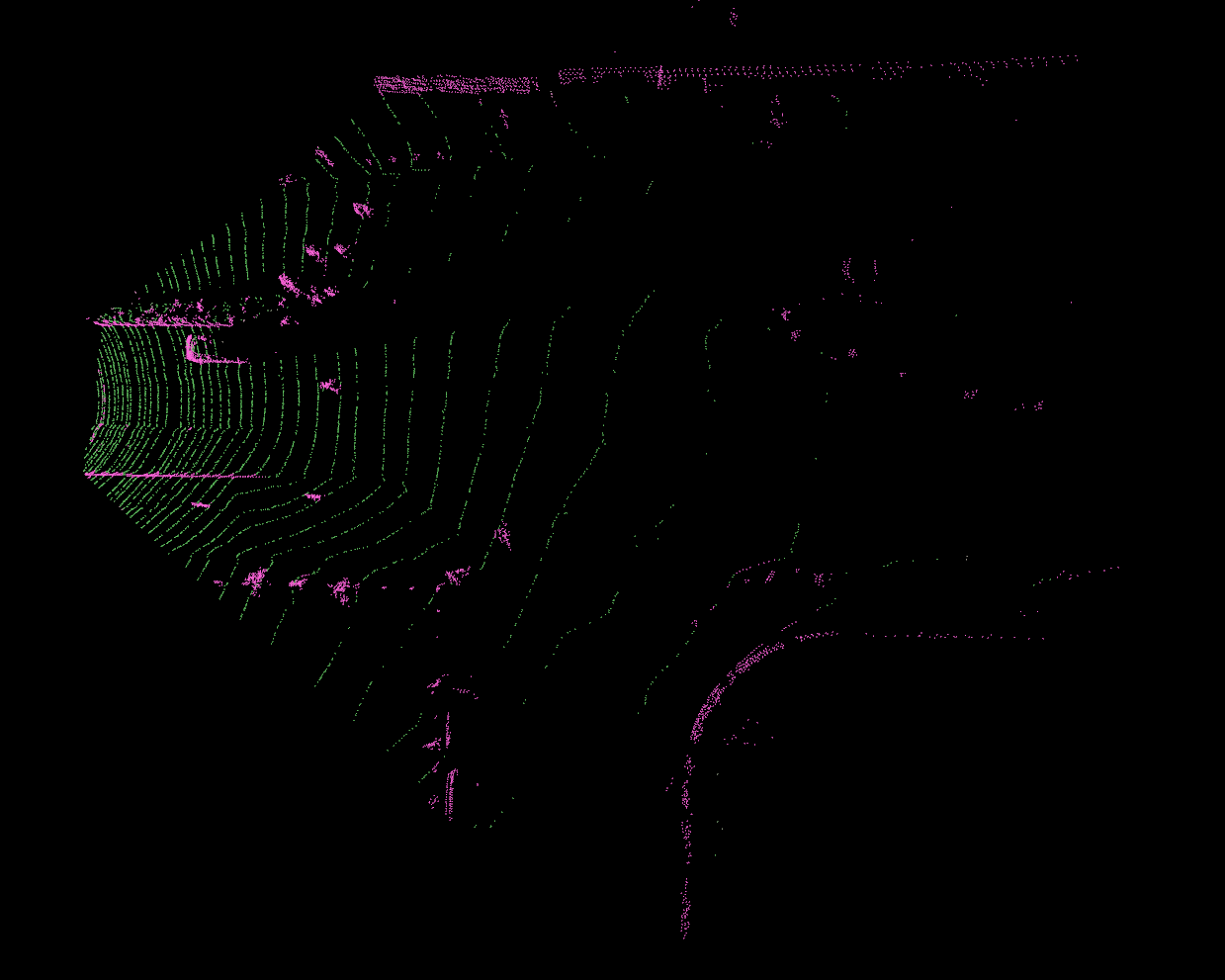}
                \caption{}
        \end{subfigure}
        \hfill
        \begin{subfigure}[b]{0.48\textwidth}
                \centering
                \includegraphics[width=\linewidth]{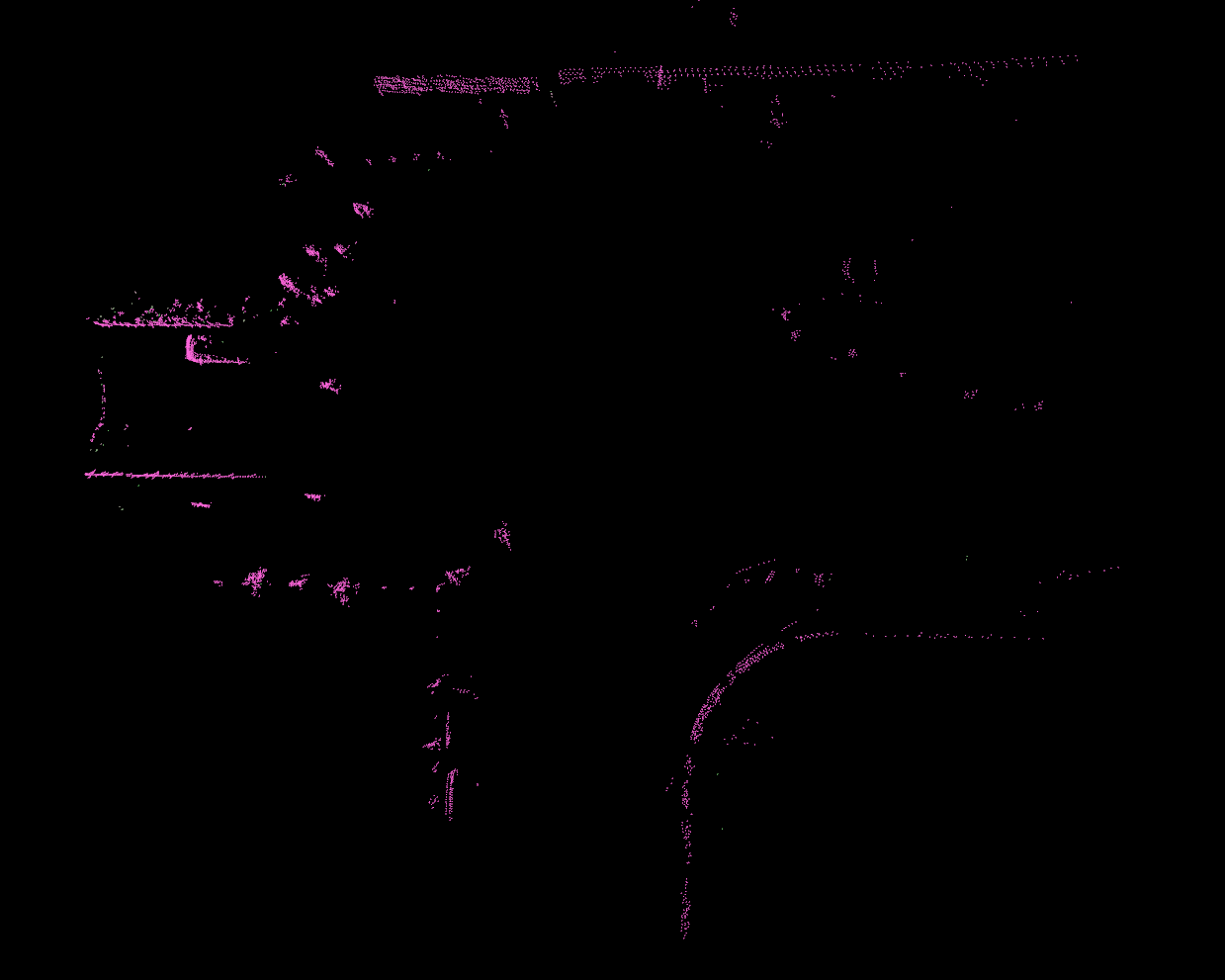}
                \caption{}
        \end{subfigure}
        \caption{Visualization of voxels of the same frame with varying drop rate, considering only normalized normal vector densities greater than 0.7: (a) original plot (b) 50\% drop rate (c) 100\% drop rate.}
        \label{fig:vis_norm_drop}
\end{figure*}

Furthermore, we can see the difference in Fig.~\ref{fig:vis_norm_drop} in the number of voxels among the three sampling percentages: no dropping, 50\%, and 100\%.
This result indicates that for some LiDAR frames, we can extract almost-pure voxels that represent high features by removing the flat plane.
Nevertheless, to avoid some incidents that may arise, we strictly drop voxels only if their normal vector densities are greater than 0.7 and drop for only 50\%.

\begin{figure*}[!ht]
        \centering
        \begin{overpic}[width=0.8\linewidth]{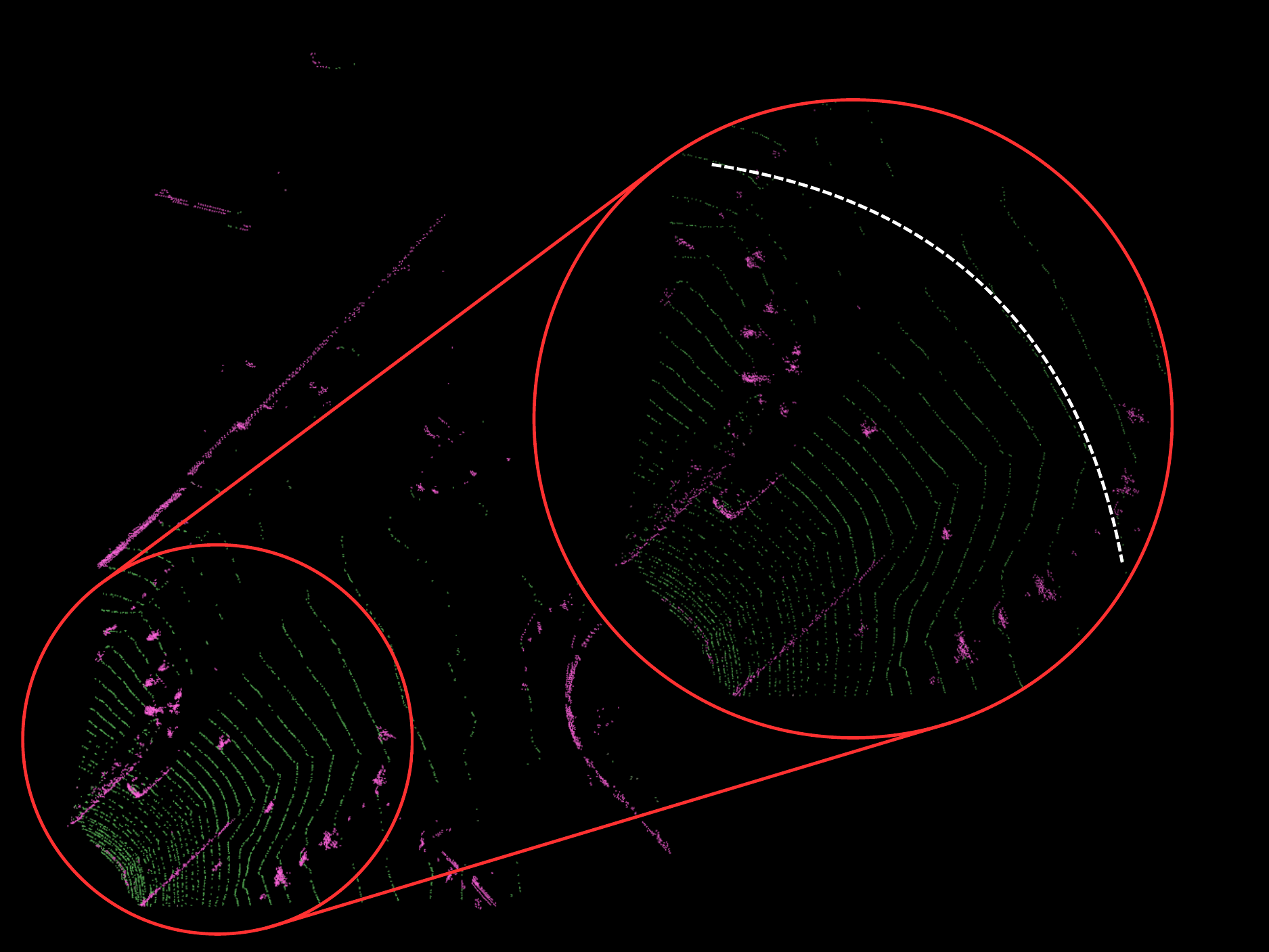}
            \put(60,21){\fontsize{9pt}{12pt}\selectfont\textcolor{white}{\(1^{st} - 4^{th}\) bin}}  
            \put(65,61.5){\fontsize{9pt}{12pt}\selectfont\textcolor{white}{\(> 30 \, \mathrm{m}\)}}  
        \end{overpic}
        \caption{Visualization of normal vector density-based sampling followed by FOV-aware bin-based sampling showing the continuous density of voxel feature.}
        \label{fig:two_samplings}
    \end{figure*}

\begin{figure}[!ht]
        \centering
        \begin{subfigure}[b]{0.48\linewidth}
                \centering
                \includegraphics[width=\linewidth]{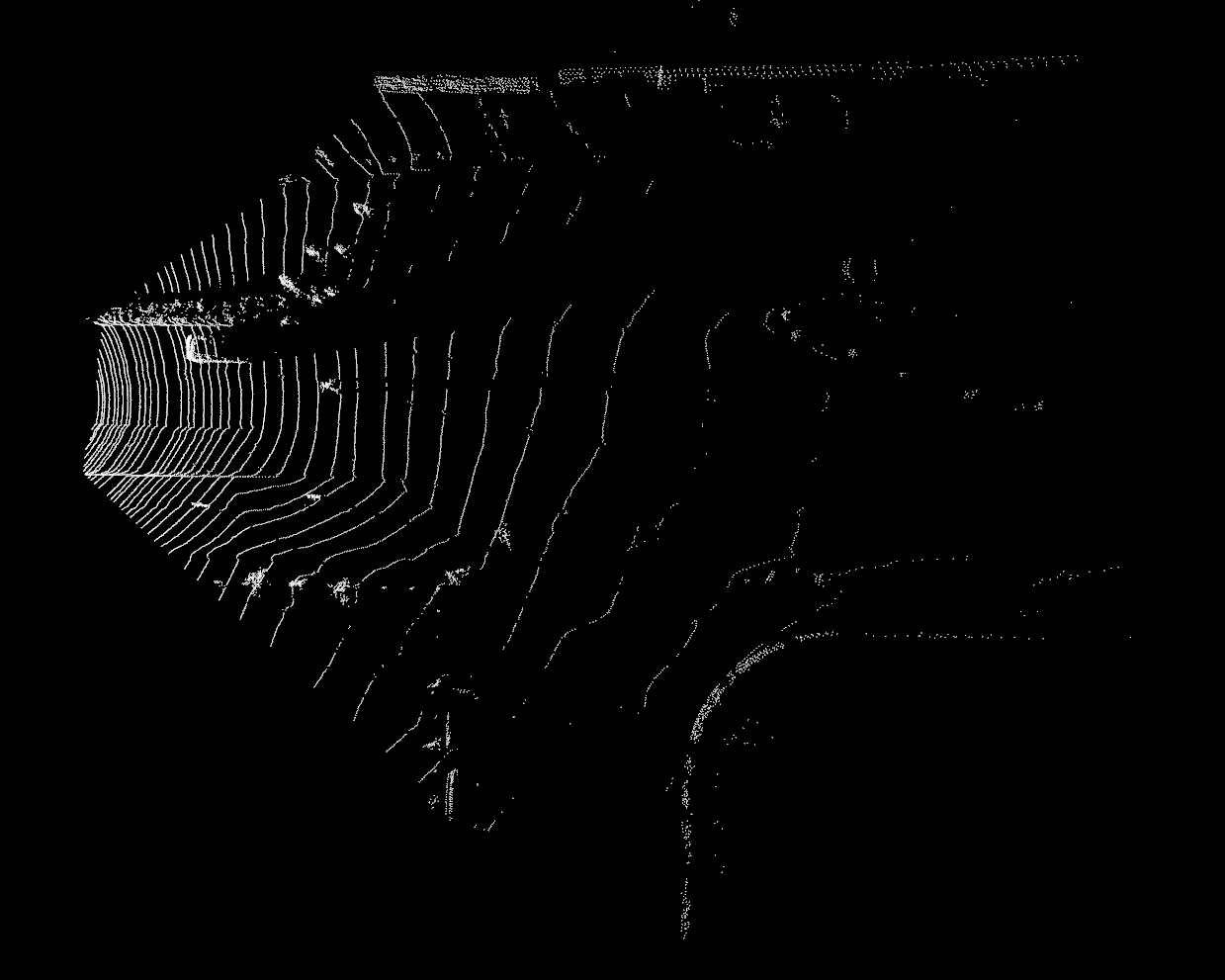}
                \caption{}
        \end{subfigure}
        \hfill
        \begin{subfigure}[b]{0.48\linewidth}
                \centering
                \includegraphics[width=\linewidth]{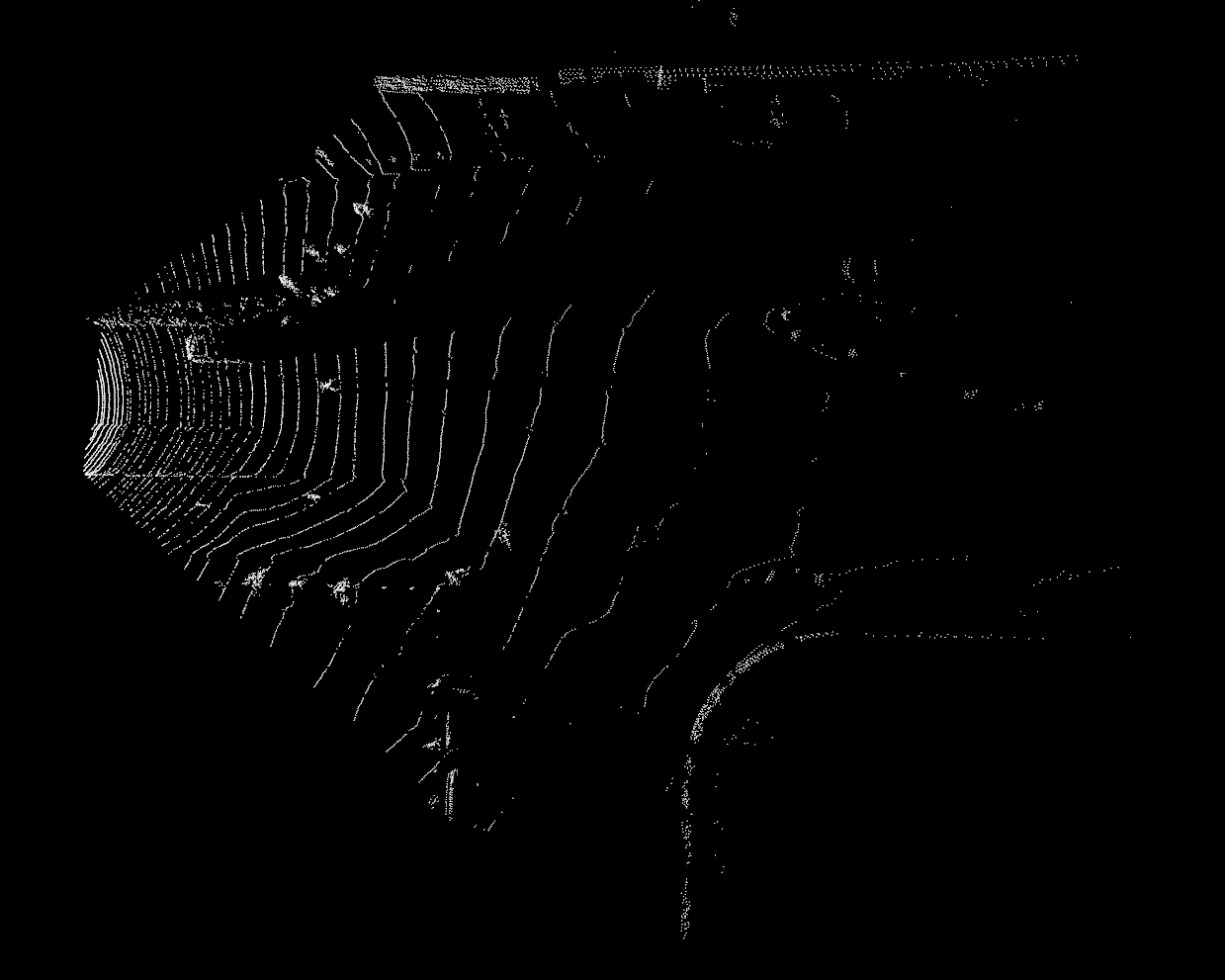}
                \caption{}
        \end{subfigure}
        \hfill
        \begin{subfigure}[b]{0.48\linewidth}
                \centering
                \includegraphics[width=\linewidth]{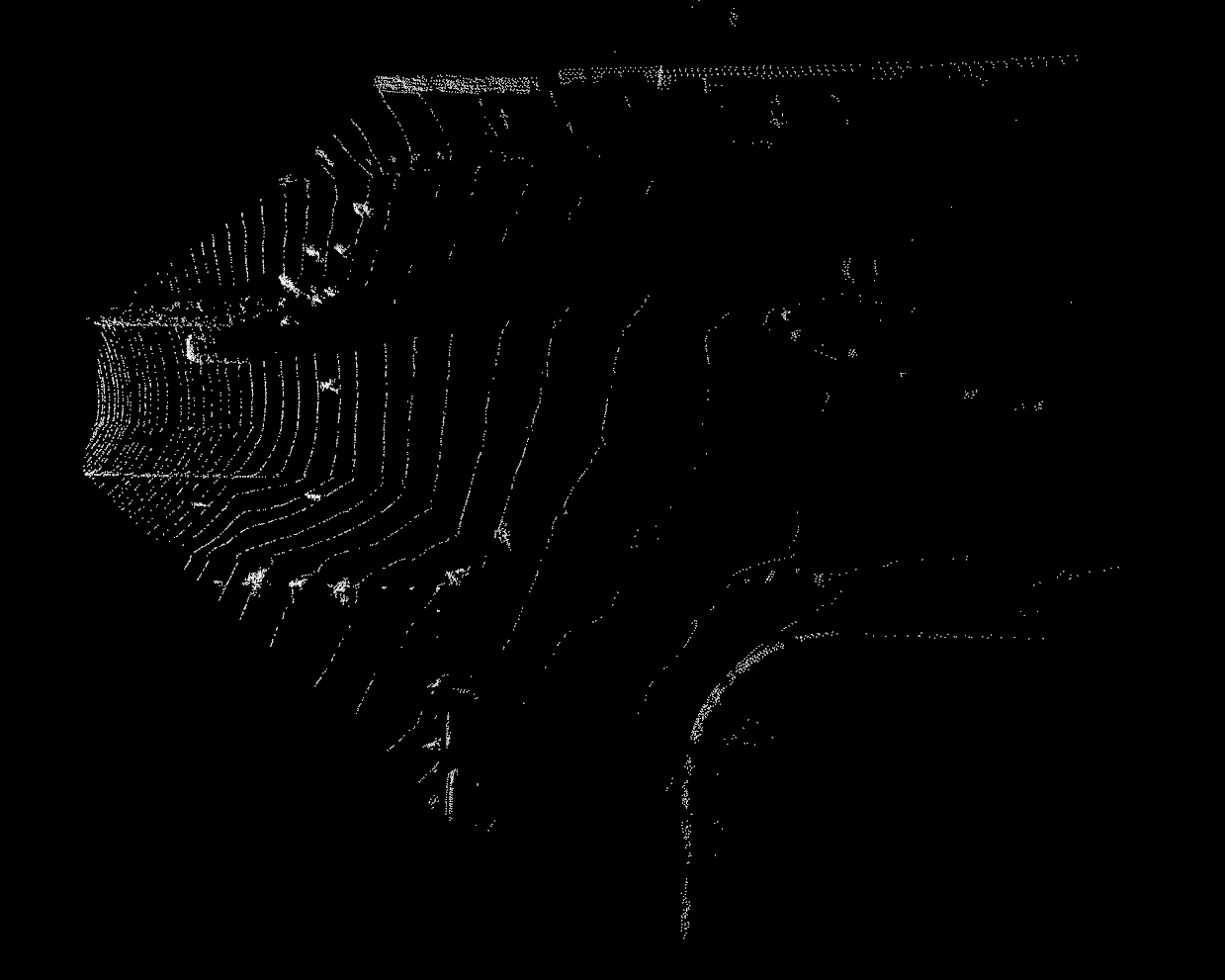}
                \caption{}
        \end{subfigure}
        \caption{Comparison of the same frame but with different sampling strategies: (a) original plot, (b) general bin-based sampling, (c) 50\% drop for points having normal vector density more than 0.7 followed by fov-aware bin-based sampling.}
        \label{fig:compare_sampling}
\end{figure}

After normal vector density-based and FOV-aware bin-based sampling are applied, voxel feature density becomes more uniformly distributed, as illustrated in Fig.~\ref{fig:two_samplings}.
This demonstrates the effectiveness of our approach in reducing uneven feature concentration, particularly in flat and redundant regions.
Furthermore, Fig.~\ref{fig:compare_sampling} presents a comparison of the original plot, general bin-based sampling, and our proposed sampling techniques, highlighting the improvements in voxel feature distribution through our proposed samplings.

\subsection{Element-Wise Attention Fusion}

\begin{figure}[!ht]
        \centering
        \includegraphics[width=0.5\linewidth]{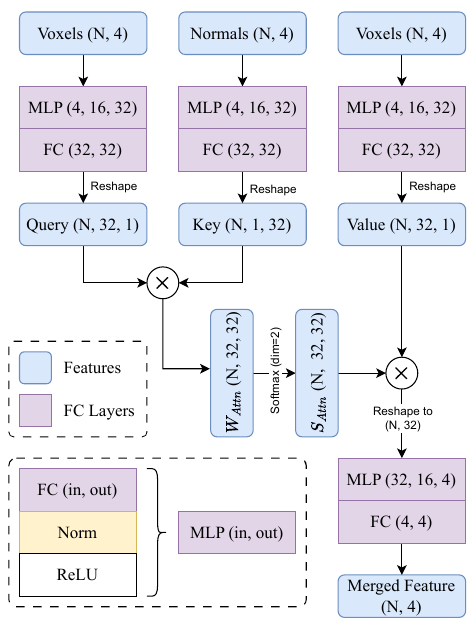}
        \caption{Illustration of element-wise attention: It follows the general attention mechanism~\cite{attn}, but with additional input features that are weighted through the encoder before fusion and then decoded as input to the 3D backbone.}
        \label{fig:norm_voxel}
\end{figure}

After normal vectors and sampling masks are obtained, both voxel  and normal vector features are then combined in the element-wise attention fusion module.
As shown in Fig.~\ref{fig:norm_voxel}, the module is implemented by processing voxel features in two different branches and normal vectors with MLP layers.
The value, query, and key are extracted, reshaped into a compatible dimension for element-wise attention, and performed another MLP layers as a decoder.
The result of this fusion module is the merged features between voxel  and normal vector features by scoring the significant feature from softmax and decoding.
Subsequently, the aggregated feature is fed into the 3D backbone for further processing.

\section{Experimental Results}

\subsection{Datasets and Evaluation Metrics}

The KITTI 3D object detection dataset~\cite{kitti} consists of 7481 and 7518 frames as LiDAR point clouds for training and testing, respectively.
Following Voxel R-CNN~\cite{voxel_rcnn} training and validating splitting strategies, we use 3712 frames for training and 3769 frames for validating results.
Also, we use 40 recall positions (R40) for evaluating average precision (AP), as suggested by~\cite{r40}.

\subsection{Implementation Details}

\textbf{Processing inputs} Following Voxel R-CNN~\cite{voxel_rcnn} input configuration, since we only consider the object within the FOV, we clip the LiDAR point clouds into [0, 70.4 m] for the X axis, [-40 m, 40 m] for the Y axis, and [-3 m, 1 m] for the Z axis.
In addition, regular voxels are derived from the raw LiDAR point clouds with sizes of (0.05 m, 0.05 m, 0.1 m), and voxel features are the mean value of point clouds for each voxel grid.
Each voxel feature is processed in the normal vector extraction module to find their neighbors for normal vector computation and sampling mask. 

\textbf{Network architecture} Element-wise attention fusion is the main processing part of NV3D.
Its encoder processes sampled voxel  and normal vector features with MLP layers of 4, 16, 32, and 32, as terms of query from normal vector features, and key and value from voxel features.
The proposed values from element-wise attention are processed in its decoder with MLP layers of 32, 32, 16, and 4.
For both the 3D and 2D backbones, we follow the Voxel R-CNN~\cite{voxel_rcnn} setup.
After processing in 3D backbone and 2D, we select the feature from the second, third, and fourth layers of the 3D backbone to generate region of interest proposals for the detection.

\textbf{Training Configuration} We use the same setup as proposed in Voxel R-CNN~\cite{voxel_rcnn}.

\subsection{Comparisons}

We first evaluate the performance of NV3D with the baseline Voxel R-CNN~\cite{voxel_rcnn} on the KITTI \textit{val} set, as shown in Table~\ref{table:nv3d_vs_baseline}.
Without any sampling technique involved, for car and cyclist, their 3D detection accuracies have improved by \textit{2.62\%} and \textit{4.23\%} mAP, respectively.
Their BEV detection accuracies have improved by \textit{3.22\%} and \textit{3.62\%} mAP, respectively.
While with both samplings, only car detection has imporved by \textit{1.56\%} \(\text{mAP}_{\text{3D}}\) and \textit{3.02\%} \(\text{mAP}_{\text{BEV}}\).

\begin{table}[!ht]
\caption{
Performance of NV3D on the KITTI \textit{val} set. These results are evaluated with the 3D and BEV AP calculated by 40 recall positions, and compared to the baseline Voxel R-CNN~\cite{voxel_rcnn}.
(a) Without sampling.
(b) With our proposed samplings.
}    
\begin{center}
\begin{subtable}[t]{0.8\textwidth}
\centering
\resizebox{\textwidth}{!}{
\begin{tabular}{c|c|cccc|cccc}
\hline
\multirow{2}{*}{Class} & \multirow{2}{*}{Method} & \multicolumn{4}{c|}{\(\text{AP}_{\text{3D}}\)} & \multicolumn{4}{c}{\(\text{AP}_{\text{BEV}}\)} \\

& & Easy & Mod. & Hard & mAP & Easy & Mod. & Hard & mAP \\
\hline
\hline

\multirow{3}{*}{Car}            & Baseline    & 89.39 & 83.83 & 78.73 & 83.98 & 90.26 & 88.35 & 87.81 & 88.81 \\
                                & NV3D   & 92.46 & 84.78 & 82.55 & 86.60 & 96.02 & 91.15 & 88.93 & 92.03 \\
                                & Diff  & \textit{+3.07} & \textit{+0.95} & \textit{+3.82} & \textit{+2.62} & \textit{+5.76} & \textit{+2.80} & \textit{+1.12} & \textit{+3.22} \\
\hline
\multirow{3}{*}{Pedestrian}     & Baseline    & 70.55 & 62.92 & 57.35 & 63.60 & 71.62 & 64.95 & 61.11 & 65.89 \\
                                & NV3D   & 64.50 & 59.60 & 54.62  & 59.57 & 68.13 & 62.31 & 58.11 & 62.85\\
                                & Diff  & \textit{-6.05} & \textit{-3.32} & \textit{-2.73} & \textit{-4.03} & \textit{-3.49} & \textit{-2.64} & \textit{-3.00} & \textit{-3.04} \\
\hline
\multirow{3}{*}{Cyclist}        & Baseline    & 90.04 & 71.13 & 66.67 & 75.95 & 91.71 & 74.67 & 70.02 & 78.80 \\
                                & NV3D   & 93.31 & 75.97 & 71.25 & 80.18 & 95.12 & 77.86 & 74.28 & 82.42 \\
                                & Diff  & \textit{+3.27} & \textit{+4.84} & \textit{+4.58} & \textit{+4.23} & \textit{+3.41} & \textit{+3.19} & \textit{+4.26} & \textit{+3.62} \\
\hline
\end{tabular}
}
\caption{}
\end{subtable}
\hfill 
\begin{subtable}[t]{\textwidth}
\centering
\resizebox{0.8\textwidth}{!}{
\begin{tabular}{c|c|cccc|cccc}
\hline
\multirow{2}{*}{Class} & \multirow{2}{*}{Method} & \multicolumn{4}{c|}{\(\text{AP}_{\text{3D}}\)} & \multicolumn{4}{c}{\(\text{AP}_{\text{BEV}}\)} \\

& & Easy & Mod. & Hard & mAP & Easy & Mod. & Hard & mAP\\
\hline
\hline

\multirow{3}{*}{Car}            & Baseline    & 89.39 & 83.83 & 78.73 & 83.98 & 90.26 & 88.35 & 87.81 & 88.81 \\
                                & NV3D   & 91.87 & 82.82 & 81.93 & 85.54 & 95.76 & 90.98 & 88.75 & 91.83 \\
                                & Diff  & \textit{+2.48} & \textit{-1.01} & \textit{+3.20} & \textit{+1.56} & \textit{+5.50} & \textit{+2.63} & \textit{+0.94} & \textit{+3.02} \\
\hline
\multirow{3}{*}{Pedestrian}     & Baseline    & 70.55 & 62.92 & 57.35 & 63.60 & 71.62 & 64.95 & 61.11 & 65.89 \\
                                & NV3D   & 49.80 & 44.85 & 41.66 & 45.44 & 53.16 & 48.82 & 45.86 & 49.28 \\
                                & Diff  & \textit{-20.75} & \textit{-18.07} & \textit{-15.69} & \textit{-18.16} & \textit{-18.46} & \textit{-16.13} & \textit{-15.25} & \textit{-16.61} \\
\hline
\multirow{3}{*}{Cyclist}        & Baseline    & 90.04 & 71.13 & 66.67 & 75.95 & 91.71 & 74.67 & 70.02 & 78.80 \\
                                & NV3D   & 87.12 & 71.57 & 67.12 & 75.27 & 91.05 & 75.23 & 70.56 & 78.95 \\
                                & Diff  & \textit{-2.92} & \textit{+0.44} & \textit{+0.45} & \textit{-0.68} & \textit{-0.66} & \textit{+0.56} & \textit{+0.54} & \textit{+0.15} \\
\hline
\end{tabular}
}
\caption{}
\end{subtable}
\end{center}        
\captionsetup{width=\textwidth}
 
\label{table:nv3d_vs_baseline}
\end{table}

Moreover, we also compare with state-of-the-art methods: CB-SSD~\cite{cb_ssd} and PVP-SSD~\cite{pvp_ssd}, as illustrated in Fig.~\ref{table:nv3d_vs_cb_ssd} and~\ref{table:nv3d_vs_pvp_ssd}.


\begin{table}[!ht]
\caption{
Performance of NV3D on the KITTI \textit{val} set. These results are evaluated with the 3D and BEV AP calculated by 40 recall positions, and compared to CB-SSD~\cite{cb_ssd}.
(a) Without sampling.
(b) With our proposed samplings.
}    
\begin{center}
\begin{subtable}[t]{0.8\textwidth}
\centering
\resizebox{\textwidth}{!}{
\begin{tabular}{c|c|cccc|cccc}
\hline
\multirow{2}{*}{Class} & \multirow{2}{*}{Method} & \multicolumn{4}{c|}{\(\text{AP}_{\text{3D}}\)} & \multicolumn{4}{c}{\(\text{AP}_{\text{BEV}}\)} \\

& & Easy & Mod. & Hard & mAP & Easy & Mod. & Hard & mAP \\
\hline
\hline

\multirow{3}{*}{Car}            & CB-SSD~\cite{cb_ssd}    & 91.28 & 82.97 & 79.81 & 84.47 & 93.45 & 89.69 & 86.86 & 90.00 \\
                                & NV3D  & 92.46 & 84.78 & 82.55 & 86.60 & 96.02 & 91.15 & 88.93 & 92.03 \\
                                & Diff  & \textit{+1.18} & \textit{+1.81} & \textit{+2.74} & \textit{+2.13} & \textit{+2.57} & \textit{+1.46} & \textit{+2.07} & \textit{+2.03} \\
\hline
\multirow{3}{*}{Pedestrian}     & CB-SSD~\cite{cb_ssd}    & 62.96 & 58.50 & 53.69 & 58.38 & 67.59 & 62.88 & 57.96 & 62.81 \\
                                & NV3D  & 64.50 & 59.60 & 54.62 & 59.57 & 68.13 & 62.31 & 58.11 & 62.85 \\
                                & Diff  & \textit{+1.54} & \textit{+1.10} & \textit{+0.93} & \textit{+1.19} & \textit{+0.54} & \textit{-0.57} & \textit{+0.15} & \textit{+0.04} \\
\hline
\multirow{3}{*}{Cyclist}        & CB-SSD~\cite{cb_ssd}    & 93.74 & 73.02 & 68.48 & 78.41 & 94.74 & 74.67 & 71.49 & 80.30 \\
                                & NV3D   & 93.31 & 75.97 & 71.25 & 80.18 & 95.12 & 77.86 & 74.28 & 82.42 \\
                                & Diff  & \textit{-0.43} & \textit{+2.95} & \textit{+2.77} & \textit{+1.77} & \textit{+0.38} & \textit{+3.19} & \textit{+2.79} & \textit{+2.12} \\
\hline
\end{tabular}
}
\caption{}
\end{subtable}
\hfill 
\begin{subtable}[t]{\textwidth}
\centering
\resizebox{0.8\textwidth}{!}{
\begin{tabular}{c|c|cccc|cccc}
\hline
\multirow{2}{*}{Class} & \multirow{2}{*}{Method} & \multicolumn{4}{c|}{\(\text{AP}_{\text{3D}}\)} & \multicolumn{4}{c}{\(\text{AP}_{\text{BEV}}\)} \\

& & Easy & Mod. & Hard & mAP & Easy & Mod. & Hard & mAP\\
\hline
\hline

\multirow{3}{*}{Car}            & CB-SSD~\cite{cb_ssd}    & 91.28 & 82.97 & 79.81 & 84.47 & 93.45 & 89.69 & 86.86 & 90\\
                                & NV3D  & 91.87 & 82.82 & 81.93 & 85.54 & 95.76 & 90.98 & 88.75 & 91.83 \\
                                & Diff  & \textit{+0.59} & \textit{-0.15} & \textit{+2.12} & \textit{+1.07} & \textit{+2.31} & \textit{+1.29} & \textit{+1.89} & \textit{+1.83} \\
\hline
\multirow{3}{*}{Pedestrian}     & CB-SSD~\cite{cb_ssd}    & 62.96 & 58.50 & 53.69 & 58.38 & 67.59 & 62.88 & 57.96 & 62.81 \\
                                & NV3D   & 49.80 & 44.85 & 41.66 & 45.44 & 53.16 & 48.82 & 45.86 & 49.28 \\
                                & Diff  & \textit{-13.16} & \textit{-13.65} & \textit{-12.03} & \textit{-12.94} & \textit{-14.43} & \textit{-14.06} & \textit{-12.10} & \textit{-13.53} \\
\hline
\multirow{3}{*}{Cyclist}        & CB-SSD~\cite{cb_ssd}    & 93.74 & 73.02 & 68.48 & 78.41 & 94.74 & 74.67 & 71.49 & 80.30 \\
                                & NV3D   & 87.12 & 71.57 & 67.12 & 75.27 & 91.05 & 75.23 & 70.56 & 78.95 \\
                                & Diff  & \textit{-6.62} & \textit{-1.45} & \textit{-1.36} & \textit{-3.14} & \textit{-3.69} & \textit{+0.56} & \textit{-0.93} & \textit{-1.35} \\
\hline
\end{tabular}
}
\caption{}
\end{subtable}
\end{center}        
\captionsetup{width=\textwidth}
\label{table:nv3d_vs_cb_ssd}
\end{table}

\begin{table}[!ht]
\caption{
Performance of NV3D on the KITTI \textit{val} set. These results are evaluated with the 3D and BEV AP calculated by 40 recall positions, and compared to PVP-SSD~\cite{pvp_ssd}.
(a) Without sampling.
(b) With our proposed samplings.
}    
\begin{center}
\begin{subtable}[t]{0.8\textwidth}
\centering
\resizebox{\textwidth}{!}{
\begin{tabular}{c|c|cccc|cccc}
\hline
\multirow{2}{*}{Class} & \multirow{2}{*}{Method} & \multicolumn{4}{c|}{\(\text{AP}_{\text{3D}}\)} & \multicolumn{4}{c}{\(\text{AP}_{\text{BEV}}\)} \\

& & Easy & Mod. & Hard & mAP & Easy & Mod. & Hard & mAP \\
\hline
\hline

\multirow{3}{*}{Car}            & PVP-SSD~\cite{pvp_ssd}   & 90.83 & 82.75 & 79.82 & 84.47 & 93.11 & 89.48 & 88.46 & 90.35 \\
                                & NV3D   & 92.46 & 84.78 & 82.55 & 86.60 & 96.02 & 91.15 & 88.93 & 92.03 \\
                                & Diff  & \textit{+1.63} & \textit{+2.03} & \textit{+2.73} & \textit{+2.13} & \textit{+2.91} & \textit{+1.67} & \textit{+0.47} & \textit{+1.68} \\
\hline
\multirow{3}{*}{Pedestrian}     & PVP-SSD~\cite{pvp_ssd}    & 70.36 & 62.65 & 56.61 & 63.20 & 73.86 & 66.38 & 61.24 & 67.16 \\
                                & NV3D   & 64.50 & 59.60 & 54.62 & 59.57 & 68.13 & 62.31 & 58.11 & 62.85 \\
                                & Diff  & \textit{-5.86} & \textit{-3.05} & \textit{-1.99} & \textit{-3.63} & \textit{-5.73} & \textit{-4.07} & \textit{-3.13} & \textit{-4.31} \\
\hline
\multirow{3}{*}{Cyclist}        & PVP-SSD~\cite{pvp_ssd}    & 92.80 & 73.40 & 68.98 & 78.39 & 95.09 & 74.74 & 71.18 & 80.34 \\
                                & NV3D   & 93.31 & 75.97 & 71.25 & 80.18 & 95.12 & 77.86 & 74.28 & 82.42 \\
                                & Diff  & \textit{+0.51} & \textit{+2.57} & \textit{+2.27} & \textit{+1.79} & \textit{+0.03} & \textit{+3.12} & \textit{+3.10} & \textit{+2.08} \\
\hline
\end{tabular}
}
\caption{}
\end{subtable}
\hfill 
\begin{subtable}[t]{\textwidth}
\centering
\resizebox{0.8\textwidth}{!}{
\begin{tabular}{c|c|cccc|cccc}
\hline
\multirow{2}{*}{Class} & \multirow{2}{*}{Method} & \multicolumn{4}{c|}{\(\text{AP}_{\text{3D}}\)} & \multicolumn{4}{c}{\(\text{AP}_{\text{BEV}}\)} \\

& & Easy & Mod. & Hard & mAP & Easy & Mod. & Hard & mAP\\
\hline
\hline

\multirow{3}{*}{Car}            & PVP-SSD~\cite{pvp_ssd}   & 90.83 & 82.75 & 79.82 & 84.47 & 93.11 & 89.48 & 88.46 & 90.35 \\
                                & NV3D   & 91.87 & 82.82 & 81.93 & 85.54 & 95.76 & 90.98 & 88.75 & 91.83 \\
                                & Diff  & \textit{+2.48} & \textit{-1.01} & \textit{+3.20} & \textit{+1.07} & \textit{+2.65} & \textit{+1.50} & \textit{+0.29} & \textit{+1.48} \\
\hline
\multirow{3}{*}{Pedestrian}     & PVP-SSD~\cite{pvp_ssd}    & 70.36 & 62.65 & 56.61 & 63.20 & 73.86 & 66.38 & 61.24 & 67.16 \\
                                & NV3D   & 49.80 & 44.85 & 41.66 & 45.44 & 53.16 & 48.82 & 45.86 & 49.28 \\
                                & Diff  & \textit{-20.56} & \textit{-17.80} & \textit{-14.95} & \textit{-17.76} & \textit{-20.70} & \textit{-17.56} & \textit{-15.38} & \textit{-17.88} \\
\hline
\multirow{3}{*}{Cyclist}        & PVP-SSD~\cite{pvp_ssd}    & 92.80 & 73.40 & 68.98 & 78.39 & 95.09 & 74.74 & 71.18 & 80.34 \\
                                & NV3D   & 87.12 & 71.57 & 67.12 & 75.27 & 91.05 & 75.23 & 70.56 & 78.95 \\
                                & Diff  & \textit{-5.68} & \textit{-1.83} & \textit{-1.86} & \textit{-3.12} & \textit{-4.04} & \textit{+0.49} & \textit{-0.62} & \textit{-1.39} \\
\hline
\end{tabular}
}
\caption{}
\end{subtable}
\end{center}        
\captionsetup{width=\textwidth}
        
\label{table:nv3d_vs_pvp_ssd}
\end{table}


We also compare with other fundamental 3D object detection frameworks~\cite{second, pointpillars, pv_rcnn, voxel_rcnn} and models that utilize local features~\cite{pointrcnn, logonet, cianet, mspv3d}, see Table~\ref{table:compare_models}.
Additionally, Table~\ref{table:overall_performance} presents the overall performance of NV3D for car detection with and without our proposed samplings.

\begin{table*}[!ht]
\caption{Performance comparison with fundamental and state-of-the-art methods on the KITTI \textit{val} set. These results are evaluated with the 3D AP calculated at 40 recall positions.}
\begin{center}
\resizebox{\textwidth}{!}{
\begin{tabular}{c|ccc|ccc|ccc}
\hline
\multirow{2}{*}{Method} & \multicolumn{3}{c|}{Cars} & \multicolumn{3}{c|}{Pedestrians} & \multicolumn{3}{c}{Cyclists} \\ 
& Easy & Mod. & Hard & Easy & Mod. & Hard & Easy & Mod. & Hard \\
\hline
\hline
SECOND~\cite{second}                 & 88.61 & 78.62 & 77.22 & 56.55 & 52.98 & 47.73 & 80.58 & 67.15 & 63.10\\
PointPillars~\cite{pointpillars}     & 87.75 & 78.39 & 75.18 & 57.31 & 51.46 & 46.83 & 81.57 & 63.00 & 59.04 \\
PV-RCNN~\cite{pv_rcnn}               & 92.57 & 84.83 & 82.69 & 62.75 & 54.47 & 49.94 & 89.09 & 70.37 & 65.98 \\ 
Voxel R-CNN~\cite{voxel_rcnn}        & 92.38 & 85.29 & 82.86 & - & - & - & - & - & -  \\
\hline
PointRCNN~\cite{pointrcnn}      & 88.72 & 78.61 & 77.82 & 62.72 & 53.85 & 50.25 & 86.84 & 71.62 & 65.59 \\ 
LoGoNet~\cite{logonet}          & 92.04 & 85.04 & 84.31 & 70.20 & 63.72 & 59.46 & 91.74 & 75.35 & 72.42 \\ 
CIANet~\cite{cianet}            & 89.68 & 84.89 & 79.32 & 67.95 & 62.54 & 56.91 & 87.21 & 73.78 & 70.69 \\ 
MSPV3D~\cite{mspv3d}            & 88.64 & 78.12 & 77.32 & 62.05 & 55.51 & 51.05 & 83.09 & 68.24 & 64.13 \\ 
\hline
\hline
NV3D (K = 7)               & \textbf{92.46} & \textbf{84.78} & \textbf{82.55} & \textbf{65.50} & \textbf{59.60} & \textbf{54.62} & \textbf{93.31} & \textbf{75.97} & \textbf{71.25} \\
\hline
\end{tabular}
}
\end{center}
\label{table:compare_models}
\end{table*}

\begin{table}[!ht]
\begin{center}
\caption{Overall performance of the NV3D fundamental detection scores on the KITTI \textit{val} set.
These results are evaluated with the AP calculated by 11 and 40 recall positions.
        (a) Without sampling.
(b) With our proposed samplings.
}        
\begin{subtable}[t]{0.49\textwidth}
\centering
\resizebox{\textwidth}{!}{
\begin{tabular}{c|ccc|ccc}
\hline
\multirow{2}{*}{Method} & \multicolumn{3}{c|}{Car mAP (R11)} & \multicolumn{3}{c}{Car mAP (R40)}\\ 
& Easy & Mod. & Hard & Easy & Mod. & Hard \\ 
\hline
\hline
BBox & 98.06 & 89.69 & 89.30 & 99.00 & 94.82 & 94.37 \\ 
BEV  & 90.36 & 88.34 & 87.83 & 96.02 & 91.15 & 88.93 \\ 
3D   & 89.32 & 84.13 & 78.77 & 92.46 & 84.78 & 82.55 \\ 
AOS  & 98.00 & 89.58 & 89.11 & 98.96 & 94.66 & 94.13 \\
\hline
\end{tabular}
}
\caption{}
\end{subtable}
\hfill
\begin{subtable}[t]{0.49\textwidth}
\centering
\resizebox{\textwidth}{!}{
\begin{tabular}{c|ccc|ccc}
\hline
\multirow{2}{*}{Method} & \multicolumn{3}{c|}{Car mAP (R11)} & \multicolumn{3}{c}{Car mAP (R40)}\\ 
& Easy & Mod. & Hard & Easy & Mod. & Hard \\ 
\hline
\hline
BBox & 97.13 & 89.67 & 89.23 & 98.81 & 94.71 & 94.14 \\ 
BEV  & 90.34 & 88.27 & 87.72 & 95.76 & 90.98 & 88.75 \\ 
3D   & 88.98 & 79.12 & 78.38 & 91.87 & 82.82 & 81.93 \\ 
AOS  & 97.02 & 89.52 & 89.01 & 98.71 & 94.51 & 93.85 \\
\hline
\end{tabular}
}
\caption{}
\end{subtable}
\captionsetup{width=\textwidth}
\label{table:overall_performance}
\end{center}
\end{table}

\section{Ablation Studies}

Our proposed sampling strategies: normal vector density-based and FOV-aware bin-based sampling, are simple tools that can be plug-in into any existing deep learning models.
In this section, we show the performance of existing deep networks that include our sampling.



\textbf{Effect of K during normal vector extracting} We investigated NV3D validation performance for difference values of \(K\) in KNN.
As shown in Table~\ref{table:nv3d_knn}, the results indicate that NV3D can achieve the highest mean average precision (mAP) of 75.56\% at \(K=7\).
Therefore, we use \(K=7\) as the default value throughout this paper.
Since KNN is performed using parallel computing, we encountered limitations due to the small number of available threads, which made the model too slow to support larger values of \(K\).

\begin{table*}[!ht]
\caption{Ablation study on NV3D trained with different \(K\) setup (\(K=3, 5, 7\))  on the KITTI \textit{val} set. These results are evaluated with the 3D AP calculated by 40 recall positions.}
\begin{center}
\resizebox{\textwidth}{!}{
\begin{tabular}{c|ccc|ccc|ccc|c}
\hline
\multirow{2}{*}{Method} & \multicolumn{3}{c|}{Car} & \multicolumn{3}{c|}{Pedestrian} & \multicolumn{3}{c|}{Cyclist}  & \multirow{2}{*}{3D mAP (R40)}\\
& Easy & Moderate & Hard & Easy & Moderate & Hard & Easy & Moderate & Hard &  \\
\hline
\hline
NV3D (K = 3)  & \textbf{92.72} & \textbf{85.17} & \textbf{82.64} & 66.25 & 60.13 & 54.52 & 90.72 & 73.34 & 68.65 & 74.90 \\
NV3D (K = 5)  & 92.61 & 84.74 & 82.53 & \textbf{67.57} & \textbf{60.46} & \textbf{55.02} & 92.36 & 72.95 & 68.67 & 75.21 \\
NV3D (K = 7)  & 92.46 & 84.78 & 82.55 & 65.50 & 59.60 & 54.62 & \textbf{93.31} & \textbf{75.97} & \textbf{71.25} & \textbf{75.56}\\
\hline
\end{tabular}
}
\end{center}
\label{table:nv3d_knn}
\end{table*}

\textbf{Effect of normal vector density-based and FOV-aware bin-based sampling} We examine two sampling techniques on the KITTI \textit{val} set with a maximum of 16000 voxels, see Table~\ref{table:eff_two_samplings}.
The result indicates that even though using full data can achieve the best result among all scenarios.
However, by applying our sampling technique, they can drop the input data up to 55\% while maintaining performance -- less than \textit{2\%} mAP drop.
Regardless, we observed that SECOND~\cite{second} yields better results after applying our samplings.

\begin{table*}[!ht]
\caption{
        Ablation study on normal vector density-based and FOV-aware bin-based sampling with the maximum number of 16,000 voxels on the KITTI \textit{val} set. These results are evaluated with the 3D AP calculated by 40 recall positions for car class.
        (ND: Normal Vector Density-based Sampling).
}
\begin{center}
\resizebox{\linewidth}{!}{
\begin{tabular}{c|c|c|ccccc}
\hline
\multirow{2}{*}{\shortstack{Sampling\\Method}} & \multirow{2}{*}{\shortstack{Drop\\Rate}} & \multirow{2}{*}{Method}  & \multicolumn{5}{c}{Cars} \\  
        & & & Easy & Medium & Hard & mAP & Diff \\ 
\hline  
\hline  
\multirow{5}{*}{No Sampling} & \multirow{5}{*}{0} 
   & SECOND~\cite{second}               & 88.61 & 78.62 & 77.22 & 81.48 & - \\
 & & PointPillar~\cite{pointpillars}    & 87.75 & 78.39 & 75.18 & 80.44 & - \\
 & & PVRCNN~\cite{pv_rcnn}              & 92.57 & 84.83 & 82.69 & 86.70 & - \\
 & & Voxel R-CNN~\cite{voxel_rcnn}      & 92.38 & 85.29 & 82.86 & 86.84 & - \\
 & & NV3D                          & 92.46 & 84.78 & 82.55 & 86.60 & - \\
\hline
\hline
\multirow{5}{*}{ND Sampling} & \multirow{5}{*}{\shortstack{\(\approx 25\)\\(24.02)}}
   & SECOND~\cite{second}               & 91.04 & 81.38 & 78.30 & 83.57 & \textit{+2.09} \\
 & & PointPillar~\cite{pointpillars}    & 87.43 & 76.24 & 73.04 & 78.90 & \textit{-1.54} \\
 & & PVRCNN~\cite{pv_rcnn}              & 92.42 & 84.66 & 82.67 & 86.58 & \textit{-0.12} \\
 & & Voxel R-CNN~\cite{voxel_rcnn}      & 92.26 & 84.83 & 82.13 & 86.41 & \textit{-0.43} \\
 & & NV3D                          & 92.40 & 83.05 & 82.35 & 85.93 & \textit{-0.67} \\
\hline
\hline
\multirow{5}{*}{FOV Sampling} & \multirow{5}{*}{\shortstack{\(\approx 50\)\\(49.16)}} 
   & SECOND~\cite{second}               & 89.70 & 80.92 & 77.59 & 82.74 & \textit{+1.26} \\
 & & PointPillar~\cite{pointpillars}    & 87.23 & 77.99 & 74.71 & 80.44 & \textit{0.00}  \\
 & & PVRCNN~\cite{pv_rcnn}              & 91.47 & 84.32 & 82.06 & 85.95 & \textit{-0.75} \\
 & & Voxel R-CNN~\cite{voxel_rcnn}      & 90.39 & 83.73 & 81.04 & 85.05 & \textit{-1.79} \\
 & & NV3D                          & 91.05 & 84.07 & 81.72 & 85.61 & \textit{-0.99} \\
\hline
\hline
\multirow{5}{*}{\shortstack{ND + FOV\\Sampling}} & \multirow{5}{*}{\shortstack{\(\approx 55\)\\(53.14)}} 
   & SECOND~\cite{second}               & 90.08 & 80.68 & 77.26 & 82.66 & \textit{+1.18} \\
 & & PointPillar~\cite{pointpillars}    & 87.61 & 76.20 & 72.92 & 78.91 & \textit{-1.53} \\
 & & PVRCNN~\cite{pv_rcnn}              & 92.27 & 83.18 & 82.65 & 86.03 & \textit{-0.67} \\
 & & Voxel R-CNN~\cite{voxel_rcnn}      & 91.38 & 84.33 & 81.71 & 85.80 & \textit{-1.04} \\
 & & NV3D                          & 91.87 & 82.82 & 81.93 & 85.54 & \textit{-1.06} \\
\hline
\end{tabular}
}
\end{center}
\label{table:eff_two_samplings}
\end{table*}



\section{Conclusion and Discussion}\label{sec12}

We implemented NV3D, a novel 3D object detection method that allows the model to use normal vector features generated from \textit{K} nearest voxel features (\(K=7\)).
NV3D utilizes neighbor features as normal vector features which help to reduce computational cost and redundancy of tensor dimension that arise from KNN.
In addition, through \textit{normal vector density-based sampling} and \textit{FOV-aware bin-based sampling}, input data can be downsampled up to 55\%.

Our proposed model and sampling methods are novel techniques that allow us to understand normal vector properties and their impact on 3D object detection.
Nevertheless, the sparsity of LiDAR point clouds should be suspected, as seen in Table~\ref{table:nv3d_vs_baseline}, NV3D performance significantly improves from its baseline Voxel R-CNN~\cite{voxel_rcnn} for cars and cyclists detection; however, the deterioration arise at detecting pedestrians.
The declined performance of pedestrian detection might happen from the behavior of human geometric shapes, as human shapes, unlike cars and cyclists that can be perceived as flat surfaces, can be round which affects the performance of normal vector estimation.
This scenario is worth more investigation, as it may be able to achieve higher performance, especially in normal vector detection techniques.









\section*{CRediT Authorship Contribution Statement}
\textbf{Krittin Chaowakarn}: Writing - original draft, Conceptualization, Methodology, Visualization, Software.
\textbf{Paramin Sangwongngam}: Writing - review, Resources, Formal analysis, Funding acquisition, Project administration.
\textbf{Nang Htet Htet Aung}: Writing - review, Formal analysis.
\textbf{Chalie Charoenlarpnopparut}: Writing - review, Resources, Supervision.

\section*{Declaration of Competing Interest}
The authors declare the following financial interests/personal relationships which may be considered as potential competing interests:
Paramin Sangwongngam reports financial support was provided by National Research Council of Thailand (NRCT). Reports a relationship with that includes:. Has patent pending to. If there are other authors, they declare that they have no known competing financial interests or personal relationships that could have appeared to influence the work reported in this paper.

\section*{Data Availability}
Data is available at \href{https://github.com/krittin-ch/NV3D}{https://github.com/krittin-ch/NV3D}.

\section*{Acknowledgements}

The first author is a scholarship recipient and has received funding support from the \textit{National Science and Technology Development Agency (NSTDA)} under the Memorandum of Understanding for the Scholarship Program for the Young Scientist and Technologist Program under a grant No. \textit{MOU-CO-2564-15054-TH}, and from the \textit{Sirindhorn International Institute of Technology (SIIT), Thammasat University (TU)}; and this work was supported in part by the \textit{National Research Council of Thailand (NRCT)} under a grant No. \textit{N42A650182}.

\bibliography{bibliography}

\end{document}